\documentclass[11pt]{article}

\PassOptionsToPackage{table}{xcolor} 
\usepackage[final]{acl}

\usepackage{times}
\usepackage{latexsym}

\usepackage[T1]{fontenc}
\usepackage[utf8]{inputenc}

\usepackage{microtype}

\usepackage{inconsolata}

\usepackage{graphicx}
\usepackage{booktabs}       
\usepackage{amsfonts}       
\usepackage{nicefrac}       

\usepackage{tcolorbox}
\tcbuselibrary{breakable, skins}

\usepackage{amsmath}
\usepackage{amssymb}
\usepackage{bbding}
\usepackage{tabularx}
\usepackage{wrapfig}
\usepackage{float}
\usepackage{placeins} 
\usepackage{enumitem}
\usepackage{colortbl}
\usepackage{makecell}
\usepackage{comment}
\usepackage{fontawesome5}

\usepackage[ruled]{algorithm2e}
\SetAlgoNoLine
\SetAlgoNoEnd
\LinesNotNumbered
\SetAlCapFnt{\footnotesize}
\SetAlCapNameFnt{\footnotesize}
\SetKwComment{Comment}{\color{gray}\# }{}
\SetAlgoNoEnd
\definecolor{lightcyan}{HTML}{CCF0FF}
\definecolor{correct}{HTML}{496DB1}
\definecolor{incorrect}{HTML}{B14C3C}
\definecolor{accblue}{HTML}{3987E5}
\definecolor{uarred}{HTML}{CF5D55}
\definecolor{trainedrow}{RGB}{224,240,248}
\definecolor{promptrow}{RGB}{235,228,247}
\newcommand{\bestscore}[1]{\textbf{#1}}

\newtcbox{\tokboxr}{%
  on line,
  colframe=Thistle,
  colback=Thistle!15,          
  coltext=Thistle!70!black,    
  boxrule=0.4pt,
  arc=1pt,
  left=0.5pt, right=0.5pt,
  top=0pt, bottom=0pt,
  enlarge left by=0pt,
  enlarge right by=0pt
}
\newtcbox{\tokboxg}{%
  on line,
  colframe=PineGreen,
  colback=PineGreen!10,        
  coltext=PineGreen!50!black,  
  boxrule=0.4pt,
  arc=1pt,
  left=1pt, right=1pt,
  top=0pt, bottom=0pt,
  enlarge left by=0pt,
  enlarge right by=0pt
}
\newtcbox{\tokboxb}{%
  on line,
  colframe=MidnightBlue,
  colback=MidnightBlue!8,      
  coltext=MidnightBlue!80!black, 
  boxrule=0.4pt,
  arc=1pt,
  left=1pt, right=1pt,
  top=0pt, bottom=0pt,
  enlarge left by=0pt,
  enlarge right by=0pt
}

\newcommand{\eg}{\textit{e.g.}, }

\usepackage{xfp}
\definecolor{accblue}{HTML}{3987E5}
\definecolor{uarred}{HTML}{CF5D55}
\newcommand{\AccLo}{13}\newcommand{\AccHi}{60}
\newcommand{\UarLo}{41}\newcommand{\UarHi}{82}
\newcommand{\Acc}[2][]{\cellcolor{accblue!\fpeval{round(5+50*min(max((#2-\AccLo)/(\AccHi-\AccLo),0),1))}}{#1#2}}
\newcommand{\Uar}[2][]{\cellcolor{uarred!\fpeval{round(5+50*min(max((#2-\UarLo)/(\UarHi-\UarLo),0),1))}}{#1#2}}

\title{
When Is Enough Not Enough? Illusory Completion in Search Agents
}

\author{
  \textbf{Dayoon Ko\textsuperscript{1}}\thanks{Work was done during internship at LG AI Research.},
  \textbf{Jihyuk Kim\textsuperscript{2}},
  \textbf{Sohyeon Kim\textsuperscript{1}},
  \textbf{Haeju Park\textsuperscript{2}},
  \textbf{Dahyun Lee\textsuperscript{2}},
\\
  \textbf{Gunhee Kim\textsuperscript{1}},
  \textbf{Moontae Lee\textsuperscript{2,3}},
  \textbf{Kyungjae Lee\textsuperscript{2,4}}\thanks{Corresponding author.}
\\
\vspace{-3mm}
\\\small{
  \textsuperscript{1}Seoul National University,
  \textsuperscript{2}LG AI Research,
  \textsuperscript{3}University of Illinois Chicago,
  \textsuperscript{4}University of Seoul
  }
\\
  \small{
  \texttt{\href{mailto:dayoon.ko@vision.snu.ac.kr}{dayoon.ko@vision.snu.ac.kr}}
  }
}

\begin{document}
\maketitle
\begin{abstract}
In agentic search, an LLM agent searches the web, reads the pages, and decides what to look for next before returning an answer.
But can we trust an answer simply because the agent returns it? Often not, and even a correct answer can be a lucky guess: on questions with several constraints, we find that in up to 48\% of their correct answers and most of their wrong ones, agents conclude the task is complete while a constraint remains unverified, which we call \emph{illusory completion}. To diagnose how it arises, we introduce the \textsc{\textbf{Epistemic Ledger}}, which tracks at every turn what has been verified and what the agent believes. Across 13 agents, training and scale increase accuracy but shift the pattern of verification failures rather than eliminating them, leaving constraints unchecked, assumed without evidence, or kept even when refuted. To probe whether explicit verification states can mitigate these failures, we simply expose each constraint's evidential state to agents at every turn using \textsc{\textbf{LiveLedger}}, a lightweight 4B tracker. With these states, accuracy improves by 4.4--16.1 points and fully verified answer rates increase by 6.2--16.7 points, suggesting that unreliable tracking of verification progress contributes to illusory completion.

\end{abstract}

\section{Introduction}

LLM-based search agents now answer complex real-world questions with growing accuracy, interleaving reasoning with search and browsing~\citep{gao2025beyond, team2025tongyi}. These questions often place several constraints on the answer (\eg a singer born in the 1990s with a Billboard Hot 100 \#1 in 2015). Thus, a reliable agent must find evidence for each one before answering. A correct final answer, however, may not show whether it did. An agent can reach a correct answer by taking shortcuts or relying on partial evidence. The same habit yields confident wrong answers when a constraint it did not check turns out not to hold.

\begin{figure}[t!]
    \includegraphics[width=1\linewidth]{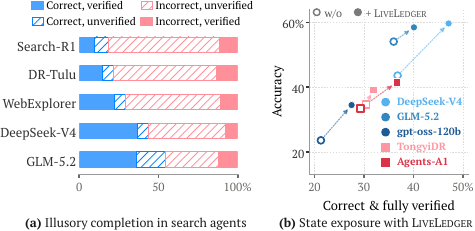} 
    \caption{\emph{Illusory completion and state exposure.} (a) Answers of five search agents, split into \textcolor{correct!80!black}{correct} and \textcolor{incorrect!70!white}{incorrect}; striped segments are unverified. Unverified answers are common, even among correct ones. (b) Accuracy against correct, fully verified answers, without (hollow) and with (filled) constraint states from \textsc{LiveLedger}. Exposing the states raises both.}
    \label{fig:teaser}
\end{figure}

Figure~\ref{fig:teaser} (a) shows that this is common. Every agent shown leaves a constraint unverified in at least 74\% of its wrong answers, and even in 16--48\% of its correct ones. We call this \emph{illusory completion}: the agent concludes that the task is complete while a constraint remains unverified. Since this failure is invisible in the final answer, we need to know how these agents reach their answers to trust them.

\begin{figure*}[t]
    \centering
    \includegraphics[width=1\linewidth]{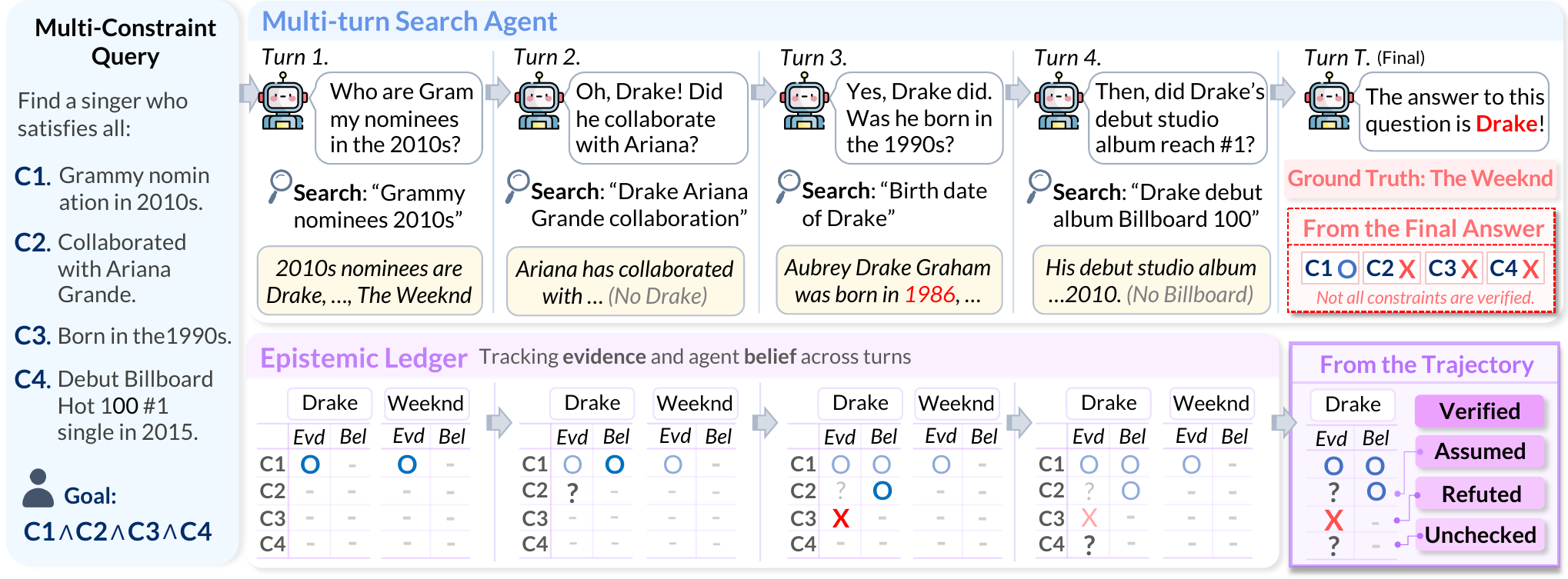}
    \vspace{-5mm}
    \caption{
    \textbf{Illusory completion through the \textsc{Epistemic Ledger}.} The agent answers Drake once $C_1$ is \textit{verified}, while its other constraints remain unverified: it affirms $C_2$ without evidence (\textit{assumed}), keeps Drake although retrieved evidence contradicts $C_3$ (\textit{refuted}), and neither establishes nor addresses $C_4$ (\textit{unchecked}). The final answer alone reveals none of these.
    }
    \label{fig:intro}
    \vspace{-2mm}
\end{figure*}

However, existing evaluations mostly judge the final answer. Some work goes further by decomposing a long-form answer to check each claim in it against retrieved sources~\citep{NEURIPS2024_27245589} or each requirement against citations in the answer~\citep{gou2025mind2web}. Even so, an answer shows only what the agent chose to report, not how it got there across its turns, where verification succeeds or fails. While \citet{xu2025beyond} examine the trajectory, they assess consistency across steps rather than progress against the task's constraints. Since verification occurs along the trajectory, against each task constraint, it should be evaluated there.

In this work, we evaluate how thoroughly search agents verify each constraint along their trajectories. Multi-constraint questions offer a clean testbed for this.
Each constraint can be checked against evidence on its own. An unverified constraint thus reflects a lapse in verification, not an error inherited from earlier reasoning steps as in multi-hop questions~\citep{trivedi2022musique}. They also demand exactly the skill at stake, since the agent must track which constraints have been verified and which remain.

To track verification progress, we introduce the \textbf{\textsc{Epistemic Ledger}}, which follows the trajectory (Figure~\ref{fig:intro}, bottom).
At each turn, for each candidate answer, it records (i) whether the evidence supports each constraint and (ii) whether the agent states that it holds.
Tracking both reveals not only whether a constraint is left unverified, but whether the agent held an unsupported belief about it or never addressed it at all.

Ideally, an agent stops only once every constraint of its committed answer is \emph{verified}, that is, established by the evidence.
An unsupported belief takes two forms: the agent \emph{assumed} the constraint without evidence ($C_2$ in Figure~\ref{fig:intro}), or kept the answer although the evidence \emph{refuted} it ($C_3$).
In all remaining cases, the constraint is \emph{unchecked}: neither established nor addressed ($C_4$).

With the \textsc{Epistemic Ledger}, we evaluate 13 agents built on models ranging from 7B RL-trained ones to frontier LLMs (Figure~\ref{fig:exp_failure_modes}).
Unchecked constraints are typically the most common failure (16--58\%), while assumed ones vary widely (2--30\%) and refuted ones stay at 3\% or below.
Two RL-trained agents on the same backbone verify a similar share of constraints but fail differently: most of one's failures are assumed, most of the other's unchecked.
Larger or stronger models leave fewer constraints unchecked, but assume more of them (2\% to 7--8\%), and refuted ones also rise (1\% to 2--3\%).
Thus, while training and scale improve accuracy, they can increase other types of failure.

To probe whether explicit verification state can reduce these failures, we expose each candidate's constraint-level progress to search agents. We use \textbf{\textsc{LiveLedger}}, a lightweight 4B tracker, as a proxy for evidential states. Without retraining the search agents, \textsc{LiveLedger} maintains an external, evidence-grounded record that keeps unresolved constraints visible throughout search. Across six agents, exposing these states improves accuracy by 4.4--16.1 points and increases fully verified answer rates by 6.2--16.7 points, with the tracker accounting for only 1.1--8.4\% of end-to-end latency. This suggests that maintaining explicit verification states can recover substantial performance during multi-turn search.

In summary, our contributions are as follows:
\begin{itemize}[leftmargin=*, topsep=2pt, itemsep=1pt, parsep=0pt, partopsep=0pt]
    \item We introduce the \textsc{Epistemic Ledger}, which tracks evidence and the agent's stated belief for each constraint along the trajectory.
    \item We identify \emph{illusory completion} and three ways a constraint is left unverified, which persist across agents, training, and model scales.
    \item Using \textsc{LiveLedger} as a proxy for evidential states, we show that exposing verification progress improves accuracy and reduces unsubstantiated answers across six trained and prompt-based agents.
\end{itemize}
\section{Related Work}
\paragraph{Beyond Final-Answer Evaluation.}
Evaluation of retrieval-augmented generation has moved beyond end-to-end accuracy, scoring the relevance of retrieved context and the faithfulness of the answer to it \citep{es-etal-2024-ragas, saad-falcon-etal-2024-ares, NEURIPS2024_27245589}.
\citet{gou2025mind2web} extend this to agentic search by checking whether the answer satisfies each task requirement and cites supporting sources, yet still judge only what the agent chose to report.
Other work evaluates the trajectory itself, checking whether reasoning, queries, and answers are consistent with one another \citep{xu2025beyond} or scoring efficiency and hallucination against accumulated evidence \citep{kim2025beyond}.
None of them tracks progress against each constraint or separates evidential support from the agent's stated belief, both of which the analysis of illusory completion requires. The \textsc{Epistemic Ledger} records evidence and stated belief for each candidate and constraint along the trajectory.

\paragraph{Multi-constraint Question Answering.}
Questions whose answer must jointly satisfy several constraints appear in knowledge-base QA and entity retrieval \citep{bao-etal-2016-constraint, talmor-berant-2018-web, malaviya-etal-2023-quest} and throughout recent benchmarks for multi-hop QA, web browsing, and deep research \citep{krishna2025fact, wolfson2026monaco, wei2025browsecomp, wu-etal-2025-webwalker}. These benchmarks, however, score only the final answer.
Some instruction-following benchmarks check each constraint separately \citep{wen2024benchmarking, he2024complex, jiang2024followbench}, but their constraints concern the model's own output. In search, by contrast, each constraint must be established from retrieved evidence. We therefore collect multi-constraint questions and evaluate how agents verify each one along the trajectory.

\paragraph{Exposing Progress to Search Agents.}
Prior work exposes progress to search agents at varying granularity: as scalar process rewards during training \citep{xu2025beyond, wen2026smartsearch, wu2025hiprag}, as binary judgments of retrieval usefulness that the agent learns to make through process rewards \citep{li2025reseek}, as rubric-based verification \citep{wan2026inference}, or as accumulated content in memory \citep{xu2026amem, zhang2026agentic}.
None makes explicit which constraint remains open for which candidate, although models revise far better when told what to fix than when left to find it themselves~\citep{lee2026refinebench}. 
Using \textsc{LiveLedger} as a proxy for evidential states, we test whether exposing constraint-level progress improves agents' answers and compare the results with process-feedback baselines.

\section{\textsc{Epistemic Ledger}}

\subsection{Search Agents}\label{sec:search_agent}

Recent search agents \citep{li2025search, jin2025searchr, gao2025beyond, liu2025webexplorer, team2025tongyi} build on the ReAct framework \citep{yao2022react}, interleaving internal reasoning with external actions such as search and browsing over multiple turns. Formally, an execution produces a trajectory $\mathcal{H}_T = (\tau_0, a_0, o_0, \ldots, \tau_T, a_T)$ \citep{team2025tongyi}, where at step $t$, $\tau_t$ is the reasoning trace, $a_t$ is the action (e.g., search, browse, or answer), and $o_t$ is the observation returned by the environment (e.g., retrieved documents). The final action $a_T$ is the answer. At each step $t \le T$, the agent samples $\tau_t, a_t \sim \pi(\cdot \mid \mathcal{H}_{t-1})$. While this loop lets the agent gather evidence over many turns, it can commit without verifying every constraint in a multi-constraint question. 

\subsection{Definition of Illusory Completion}
\label{sec:definition}
 
Given a question with constraints $\mathcal{C} = \{C_1, \ldots,$ $C_n\}$ that a valid answer must jointly satisfy, an agent returns an \emph{unsubstantiated} answer if it terminates at step $T$ with $a_T$ such that some $C_i \in \mathcal{C}$ is not established by the evidence for $a_T$. We define \textbf{illusory completion} as terminating with such an answer while treating the task as complete. Because the answer still looks plausible, this failure is invisible to final-answer evaluation and must be diagnosed along the trajectory.

\subsection{Tracking Evidence and Belief}

To diagnose illusory completion, we introduce the \textsc{Epistemic Ledger}, which jointly tracks evidential support and the agent's stated belief.
Evidential support identifies unsubstantiated answers, and together with belief, it assigns each constraint of the committed answer one of four outcomes (\S\ref{sec:illusory_completion}).
The ledger is maintained by an evaluator model $\mathcal{M}$, for which we use \texttt{gpt-5-nano} \citep{singh2025openai}.
All prompts are in Appendix~\ref{sec:appx_prompt}.

\paragraph{Overview.}
As shown in Figure~\ref{fig:intro}, the ledger records two components for each candidate answer $k \in \mathcal{L}$ (\eg Drake and The Weeknd): (i) the \textit{constraint states}, what the evidence shows and what the agent states about each constraint $C_i \in \mathcal{C}$, and (ii) the \textit{candidate status}, the agent's commitment to $k$.
Given a question and trajectory $\mathcal{H}_T$, the evaluator $\mathcal{M}$ first decomposes the question into externally verifiable constraints $\mathcal{C}=\{C_1,\ldots,C_n\}$ and initializes $\mathcal{L}_0$.
At each step $t$, it extracts candidates from $o_t$ (\eg Drake and The Weeknd from the first search) and updates their constraint states and status, yielding $\mathcal{L}_T$ after $T$ steps.
Algorithms~\ref{alg:ledger-structure} and~\ref{alg:ledger-update} give full details.

\paragraph{Constraint States.}
For each $(k, C_i)$, the ledger records evidential support $E$ and stated belief $B$,
\vspace{-1mm}
{\fontsize{10.5}{12.5}\selectfont
\begin{align}
E(k, C_i) &\in \{\textsc{Satisfied},\, \textsc{Disproved},\, \textsc{Unknown}\}, \nonumber \\
B(k, C_i) &\in \{\textsc{Affirm},\, \textsc{Deny},\, \textsc{Unaddressed}\}. \nonumber
\end{align}
}
\vspace{-6mm}

$E(k, C_i)$ reflects what the retrieved evidence shows about whether $k$ satisfies $C_i$, and $B(k, C_i)$ reflects what the agent's reasoning states about it.
In Figure~\ref{fig:intro}, $E(\text{Drake}, C_1)$ and $E(\text{The Weeknd}, C_1)$ become \textsc{Satisfied} once the first search returns supporting evidence, and $E(\text{Drake}, C_3)$ becomes \textsc{Disproved} when the search in Turn~3 reports a 1986 birth year.
$B(\text{Drake}, C_2)$ becomes \textsc{Affirm} in Turn~3 although no evidence supports it.

\paragraph{Candidate Status.}
For each candidate $k \in \mathcal{L}$, the ledger also records $\text{status}(k) \in \{\textsc{Committed},$ $\textsc{Stored}, \textsc{Rejected}\}$: \textsc{Committed} if the agent currently pursues $k$ as its working answer, \textsc{Stored} if $k$ has been mentioned but is not being pursued, remaining an alternative, and \textsc{Rejected} if the agent has explicitly ruled it out.
In Figure~\ref{fig:intro}, Drake is \textsc{Committed} throughout the trajectory, while The Weeknd remains \textsc{Stored}.
We analyze the constraints of the \textsc{Committed} candidate in $\mathcal{L}_T$. 

\paragraph{Update Rules.}
$\mathcal{M}$ updates the ledger incrementally along $\mathcal{H}_T$ in two stages per step.
Given $\mathcal{L}_{t-1}$ and the tuple $(\tau_t, a_t, o_t)$, it first identifies new candidates in $o_t$, initializes their states to $E = \textsc{Unknown}$ and $B = \textsc{Unaddressed}$, and updates $E(k, C_i)$ for all candidates from $o_t$.
It then reads the subsequent reasoning $\tau_{t+1}$ to update $B(k, C_i)$ and the candidate status.

\begin{algorithm}[t]
\caption{\textsc{Epistemic Ledger} Structure}
\label{alg:ledger-structure}
\footnotesize
\hspace{-3mm}\textbf{Constraint States:} $\mathcal{L}[k][C_i]$ for $(k, C_i)$\;
\Indp
\hspace{-8mm}$E(k, C_i) \in \{\textsc{Satisfied},\, \textsc{Disproved},\, \textsc{Unknown}\}$\;
\hspace{-8mm}$B(k, C_i) \in \{\textsc{Affirm},\, \textsc{Deny},\, \textsc{Unadd.}\}$\;
\Indm
\hspace{-3mm}\textbf{Candidate Status:} $\textnormal{status}(k)$\;
\Indp
\hspace{-8mm}$\textnormal{status}(k) \in \{\textsc{Committed},\, \textsc{Stored},\, \textsc{Rejected}\}$\;
\Indm
\end{algorithm}

\begin{algorithm}[h]
\caption{\textsc{Epistemic Ledger} Update}
\label{alg:ledger-update}
\footnotesize
\KwIn{Trajectory $\mathcal{H}_T$, Constraints $\mathcal{C}$, Evaluator $\mathcal{M}$}
\KwOut{Updated ledger $\mathcal{L}_T$}
Initialize $\mathcal{L}_0 \leftarrow \emptyset$\;
\For{$t = 0$ \KwTo $T-1$}{
    $(\tau_t, a_t, o_t, \tau_{t+1}) \leftarrow$ current/next turn\;
    \tcc{update candidates}
    $\mathcal{L}_t \leftarrow \mathcal{M}.\text{add\_candidates}(\mathcal{L}_t, o_t)$\;
    \tcc{update $E, B$}
    \ForEach{$k \in \mathcal{L}_t$, $C_i \in \mathcal{C}$}{
        \uIf{$o_t$ proves $(k, C_i)$}{
            $E(k, C_i) \leftarrow \textsc{Satisfied}$\;
        }
        \ElseIf{$o_t$ disproves $(k, C_i)$}{
            $E(k, C_i) \leftarrow \textsc{Disproved}$\;
        }
        \uIf{agent affirms $(k, C_i)$}{
            $B(k, C_i) \leftarrow \textsc{Affirm}$\;
        }
        \ElseIf{agent denies $(k, C_i)$}{
            $B(k, C_i) \leftarrow \textsc{Deny}$\;
        }
    }
    \tcc{update status}
    \ForEach{$k \in \mathcal{L}_t$}{
        \uIf{agent selects $k$}{
            $\textnormal{status}(k) \leftarrow \textsc{Committed}$\;
        }
        \uElseIf{agent abandons $k$}{
            $\textnormal{status}(k) \leftarrow \textsc{Rejected}$\;
        }
        \Else{
            $\textnormal{status}(k) \leftarrow \textsc{Stored}$\;
        }
    }
    $\mathcal{L}_{t+1} \leftarrow \mathcal{L}_t$\;
}
\Return{$\mathcal{L}_T$}\;
\end{algorithm}

\subsection{Constraint Outcomes}\label{sec:illusory_completion}\label{sec:under_verified_answer}

Using $\mathcal{L}_T$, the committed answer $a_T$ is unsubstantiated (\S\ref{sec:definition}) if $E(a_T, C_i) \neq \textsc{Satisfied}$ for some $C_i \in \mathcal{C}$, even when $a_T$ is correct. To see how verification fails, we assign each constraint an outcome $\ell(a_T, C_i)$ from its final state in $\mathcal{L}_T$, checking evidence first and belief second.

\vspace{-1mm}
\paragraph{Verified.}
The evidence establishes the constraint ($E(a_T, C_i)=\textsc{Satisfied}$), \eg (Drake, $C_1$).

\vspace{-1mm}
\paragraph{Assumed.}
No evidence settles the constraint ($E(a_T, C_i)=\textsc{Unknown}$), but the agent states it holds ($B(a_T, C_i)=\textsc{Affirm}$), \eg (Drake, $C_2$).

\vspace{-1mm}
\paragraph{Refuted.}
The evidence contradicts the constraint ($E(a_T, C_i)=\textsc{Disproved}$), yet the agent keeps $a_T$, whether it affirms the constraint or leaves it unaddressed, \eg{} (Drake, $C_3$).

\vspace{-1mm}
\paragraph{Unchecked.}
No evidence settles the constraint, and the agent never addresses it ($E(a_T, C_i)=\textsc{Unknown}$, $B(a_T, C_i)=\textsc{Unaddressed}$), \eg{} (Drake, $C_4$).

The agent rarely denies a constraint of its committed candidate (1.3\% of constraints), so we exclude these, and the four outcomes cover the rest.

%

\section{Probing Verification Failures with \textsc{LiveLedger}}\label{sec:liveledger}
\vspace{-1mm}
Can making verification progress explicit reduce illusory completion? We test this by exposing per-constraint evidential states to the same agents at every turn, using \textsc{\textbf{LiveLedger}}, an inference-time ledger updater, as a proxy.

Given a question, \textsc{LiveLedger} decomposes it into constraints $\hat{\mathcal{C}}$ and initializes $\hat{\mathcal{L}}_0$.
After each step $t$, it adds candidates found in $o_t$ and updates their evidential support $E(k, C_i)$, and the agent samples $\tau_{t+1}, a_{t+1} \sim \pi(\cdot \mid \mathcal{H}_t, \hat{\mathcal{L}}_{t+1})$.
\textsc{LiveLedger} records only evidential support, so that the state reflects what the evidence shows rather than what the agent has claimed.

The state $\hat{\mathcal{L}}_t$, with its constraint list, is the only addition to the agent's context, so any change in behavior comes from exposing it.
Unlike process-feedback methods, which compress progress into a scalar reward \citep{wu2025hiprag} or a binary usefulness judgment \citep{li2025reseek}, \textsc{LiveLedger} assesses each constraint separately and keeps it visible across turns.

We obtain \textsc{LiveLedger} by fine-tuning Qwen3.5-4B on ledger updates generated by \texttt{gpt-oss-120b}. We collect these updates on MultiConIR \citep{lu2025multiconirmulticonditioninformationretrieval} and HDS-QA \citep{ko2025hybrid}, keeping only trajectories that reach correct answers and pass a verification check. 
A small tracker keeps per-turn state exposure practical. We assess its reliability and cost in Appendices~\ref{sec:appx_validation} and~\ref{sec:appx_cost}. Implementation and training details are in Appendix~\ref{sec:appx_liveledger_details}.

\section{Experiments}\label{sec:experiments}

\subsection{Experiment Setup}

\paragraph{Datasets.}
We gather 484 multi-constraint questions from six benchmarks: BrowseComp \citep{wei2025browsecomp}, DeepSearchQA \citep{deepsearchqa2025}, FRAMES \citep{krishna2025fact}, LiveDRBench \citep{java2025characterizing}, WebWalkerQA \citep{wu-etal-2025-webwalker}, and BioASQ \citep{krithara2023bioasq}.
Following \citet{vermaplan}, we decompose each question into a directed acyclic graph (DAG) of its reasoning structure and keep those with width $\geq 3$ and depth $= 1$, that is, questions whose answer must satisfy three or more independent constraints.
We exclude deeper DAGs because a failure there may lie in resolving an intermediate entity rather than in verifying constraints.
More details are in Appendix~\ref{sec:appx_dataset}.

\paragraph{Baselines.}
We evaluate 13 agents.
Seven are trained: \textbf{Search-R1} and \textbf{ASearcher}, trained with RL for multi-turn search; \textbf{RAG-R1}, trained with GRPO to issue multiple queries per turn; \textbf{WebExplorer}, which unifies reasoning with iterative search and browsing; and three deep research agents, \textbf{DR-Tulu}, trained with evolving RL rubrics, \textbf{Tongyi DeepResearch (TongyiDR)}, trained with GRPO for long-horizon search, and \textbf{Agents-A1}.

Six are prompt-based: \textbf{Search-o1}, which summarizes retrieved documents for large reasoning models, with \texttt{gpt-oss-20b} and \texttt{gpt-oss-120b} as backbones, and four LLMs that call the search and browse tools directly in a reasoning loop: \textbf{gpt-oss-20b}, \textbf{gpt-oss-120b}, \textbf{DeepSeek-V4-Pro}, and \textbf{GLM-5.2}.
All agents can search, and all except Search-R1, RAG-R1, and Search-o1 can also browse.

We first diagnose all 13 with the \textsc{Epistemic Ledger}, and then give \textsc{LiveLedger} to six of them: TongyiDR, Agents-A1, and the four LLMs.
To compare \textsc{LiveLedger} with process feedback, we also evaluate \textbf{HiPRAG}~\cite{wu2025hiprag}, trained with process rewards against over- and under-search, and \textbf{ReSeek}~\cite{li2025reseek}, which learns to judge retrieval usefulness through process rewards, both on Qwen2.5-7B, the same backbone as Qwen2.5-7B with \textsc{LiveLedger}.

\paragraph{Metrics.}\label{sec:exp_setup}
We report two primary metrics: \textbf{(i) accuracy (Acc)}, the proportion of correct answers judged by \texttt{gpt-5.6-sol}; and \textbf{(ii) unsubstantiated answer rate (UAR)}, the share of questions whose final answer $a_T$ is unsubstantiated  (\S\ref{sec:under_verified_answer}). Specifically, an answer is fully verified when the retrieved evidence establishes all constraints, so the fully verified answer rate is the complement of UAR. In a human evaluation of 100 samples with the evaluator, majority votes of three annotators agree with UAR on 91\% of them (Fleiss' $\kappa=0.76$; more details are in Appendix~\ref{sec:uar_human_evaluation}).

To follow verification progress, we retrospectively track the final answer $a_T$ through the \textsc{Epistemic Ledger}. A run is stalled
at step $t$ if it has not yet answered, $a_T$ has gained no \textsc{Satisfied} constraint over the last three steps, and at least one of its constraints remains unverified in $L_t$. 

\begin{figure*}[t!]
    \centering
    \includegraphics[width=\linewidth]{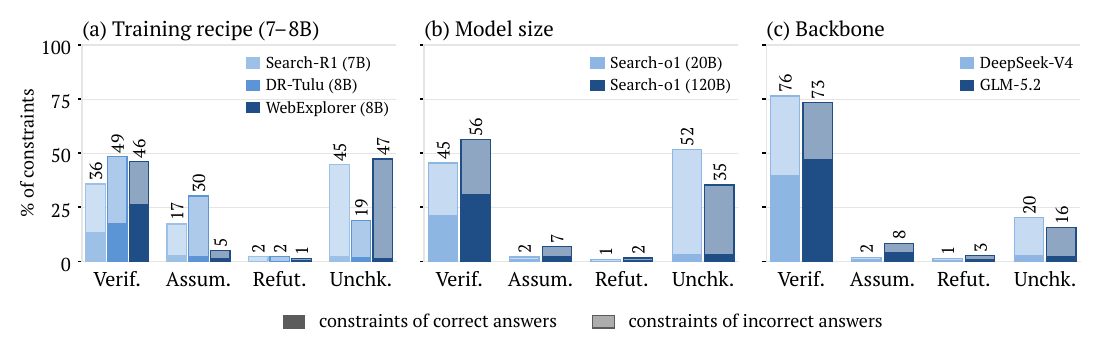}
    \vspace{-8mm}
    \caption{\textbf{Constraint outcomes of the \textsc{Committed} candidate.} For each question, we compute the share of the candidate's constraints in $\mathcal{L}_T$ that are \textit{verified}, \textit{assumed}, \textit{refuted}, or \textit{unchecked}, and average it over all questions: (a) training recipes at 7--8B, (b) model size, and (c) backbone. Each bar is split by the run result into correct (dark) and incorrect (light) answers. Higher accuracy shifts constraints between failures rather than reducing all of them.}
    \label{fig:exp_failure_modes}
\end{figure*}

\paragraph{Experimental Details.} 
We use the Serper API for web search and the Jina API for browsing (the top-10 results per search call).
Each run is capped at 30 agent turns (Appendix~\ref{sec:appx_hyperparameters}).
We run each agent once for cost reasons, but repeat three of them three times to check for consistency (Appendix~\ref{sec:appx_three_runs}). Appendix~\ref{sec:appx_cost} reports the cost of the evaluator and \textsc{LiveLedger}, and Appendix~\ref{sec:appx_qualitative_examples} shows example trajectories with their ledgers.
Further details are in Appendix~\ref{sec:appx_experiment_details}.

\subsection{Diagnosing Illusory Completion}\label{sec:exp_diagnosis}

\paragraph{Most answers are unsubstantiated, even from the strongest agents.}
Table~\ref{tab:acc-uar-results} reports Acc and UAR.
Agents with higher Acc tend to have lower UAR, yet no agent returns a substantiated answer on even half of the questions: UAR ranges from 52.1\% (GLM-5.2) to 80.4\% (ASearcher).
Among trained agents, TongyiDR achieves the highest Acc (34.7\%) while still leaving 64.0\% of questions with unsubstantiated answers. Accuracy also overstates verification: 16--48\% of correct answers are unsubstantiated (Figure~\ref{fig:teaser}).

\begin{table}[t]
\centering
\footnotesize
\setlength{\tabcolsep}{2.5pt}
\renewcommand{\arraystretch}{1.1}
\begin{tabularx}{\columnwidth}{@{}Xcc@{}}
\toprule
Agent & Acc ($\uparrow$) & UAR ($\downarrow$) \\
\midrule
\multicolumn{3}{c}{\textit{Trained}} \\
\midrule
Search-R1 \citep{jin2025searchr}                  & \cellcolor{accblue!6}18.6 & \cellcolor{uarred!34}78.9 \\
ASearcher \citep{gao2025beyond}                   & \cellcolor{accblue!6}18.0 & \cellcolor{uarred!36}80.4 \\
RAG-R1 \citep{tan2025rag}                         & \cellcolor{accblue!6}18.0 & \cellcolor{uarred!35}79.3 \\
DR-Tulu \citep{shao2025dr}                        & \cellcolor{accblue!9}21.7 & \cellcolor{uarred!27}71.5 \\
WebExplorer \citep{liu2025webexplorer}            & \cellcolor{accblue!15}28.9 & \cellcolor{uarred!22}66.9 \\
Agents-A1 \citep{bai2026scalinghorizonparametersreaching} & \cellcolor{accblue!19}33.5 & \cellcolor{uarred!20}65.7 \\
TongyiDR \citep{team2025tongyi}                   & \cellcolor{accblue!20}\bestscore{34.7} & \cellcolor{uarred!19}\bestscore{64.0} \\
\addlinespace[2pt]
\midrule
\multicolumn{3}{c}{\textit{Prompt-based}} \\
\midrule
Search-o1 (20b) \citep{li2025search}              & \cellcolor{accblue!12}25.4 & \cellcolor{uarred!21}66.3 \\
Search-o1 (120b) \citep{li2025search}             & \cellcolor{accblue!21}36.2 & \cellcolor{uarred!16}61.6 \\
gpt-oss-20b \citep{agarwal2025gpt}                & \cellcolor{accblue!10}22.3 & \cellcolor{uarred!26}70.9 \\
gpt-oss-120b \citep{agarwal2025gpt}               & \cellcolor{accblue!11}23.6 & \cellcolor{uarred!26}70.7 \\
DeepSeek-V4-Pro \citep{deepseekai2026v4}          & \cellcolor{accblue!27}43.6 & \cellcolor{uarred!10}55.6 \\
GLM-5.2 \citep{glm5team2026glm5vibecodingagentic} & \cellcolor{accblue!36}\bestscore{54.1} & \cellcolor{uarred!6}\bestscore{52.1} \\
\bottomrule
\end{tabularx}
\vspace{-2mm}
\caption{
  \textbf{Main results on all baselines.}
  Acc (\%) and UAR (\%) on the 484 questions.
  Darker cells mean higher values.
  The best score in each group is in \textbf{bold}.
}
\label{tab:acc-uar-results}
\end{table}

\vspace{-1mm}
\paragraph{Training recipes shift failures between assumed and unchecked constraints.}
For each question, we compute the share of the \textsc{Committed} candidate's constraints in each outcome. Figure~\ref{fig:exp_failure_modes} reports these shares averaged over the 484 questions. Each bar is split by the run result into correct (dark) and incorrect (light) answers, so the dark parts sum to the accuracy.

In (a), DR-Tulu achieves 3.1 points higher Acc than Search-R1 and leaves 25.9 points fewer constraints unchecked. But only half of this difference is verified (+12.7); the other half is assumed (+13.0), all in incorrect answers (14.4 to 28.0 points). Meanwhile, WebExplorer, with 7.2 points higher Acc than DR-Tulu, verifies a similar share of constraints, but shows the opposite pattern: 25.2 points fewer assumed constraints but 28.5 points more unchecked ones.

Even as accuracy improves, different training recipes lead to different failures, so accuracy alone does not show how an agent fails.

\vspace{-1mm}
\paragraph{Larger or stronger models often assume more constraints.}
In (b), Search-o1 with the 120B backbone achieves 10.8 points higher Acc than with the 20B backbone. It verifies 10.8 points more constraints and leaves 16.3 points fewer unchecked, but assumes 4.8 points more. Most of the added assumed constraints are in incorrect answers (3.7 of the 4.8 points).

In (c), GLM-5.2 achieves 10.5 points higher Acc than DeepSeek-V4-Pro, yet verifies 2.9 points fewer constraints and assumes 6.5 points more.
More than half of the added assumed constraints are in correct answers (3.5 of the 6.5 points): GLM-5.2 answers correctly while affirming constraints it has not verified.
Refuted ones also increase in both cases, from 0.8 to 1.6 and from 1.4 to 2.6 points, although they remain below 3\% for every agent.
This shows that even as scale improves accuracy, agents can also assume more constraints, which a correct answer does not reveal.

\paragraph{Unchecked constraints coexist with verified ones.}
A constraint left unchecked often sits on a candidate that is already partly verified.
For ten of the agents, 65--88\% of the questions with an unchecked constraint also have a verified constraint on the same candidate; for the weaker Search-R1, ASearcher, and RAG-R1, only 35--44\% do.
Weaker agents thus rarely make progress on their candidate, while stronger agents make progress but leave a constraint unchecked.

In these questions, stronger agents verify more of the candidate's constraints (42--57\% vs.\ 36--39\%), yet only 4--24\% are answered correctly, compared with 34--68\% when every constraint is verified.
So verifying more constraints does not always raise accuracy: while one constraint remains unchecked, the added verification rarely yields a correct answer.

\begin{figure*}[t]
    \centering
    \includegraphics[width=\linewidth]{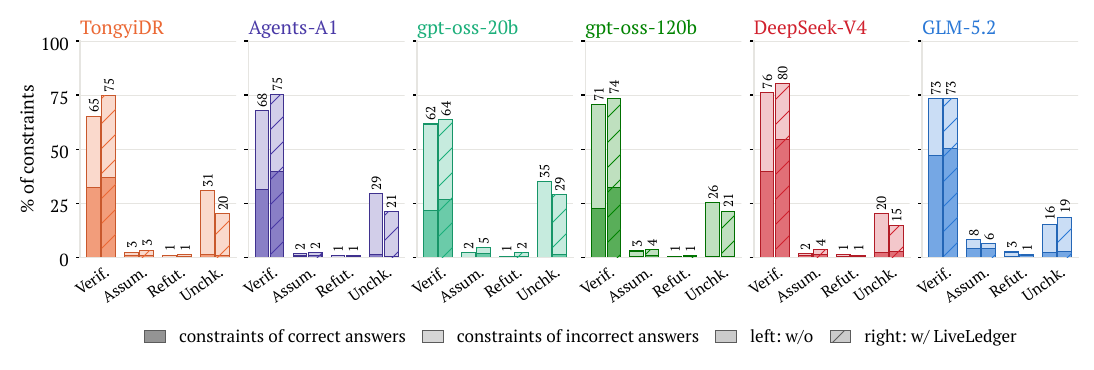}
    \vspace{-8mm}
    \caption{\textbf{Constraint outcomes with \textsc{LiveLedger}.} 
    For each agent, the left bar of each outcome is without \textsc{LiveLedger} and the hatched right bar is with it, computed and split by the run result as in Figure~\ref{fig:exp_failure_modes}. With \textsc{LiveLedger}, most agents verify more constraints and leave fewer unchecked (by up to 10.5 points), with far smaller shifts in assumed and refuted ones (at most 2.4 points); more of the verified constraints come from correct answers, raising accuracy.
    }
    \vspace{-3mm}
    \label{fig:ll_condition}
\end{figure*}

\begin{figure*}[t]
    \centering
    \includegraphics[width=\linewidth]{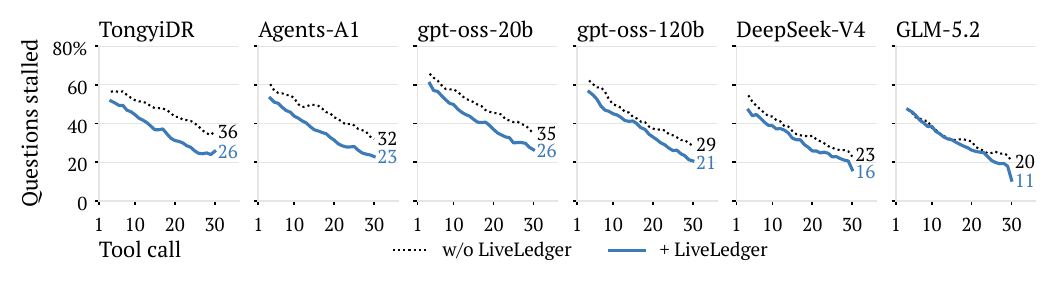}
    \vspace{-8mm}
    \caption{\textbf{Stalled runs with and without \textsc{LiveLedger}.} Share of questions stalled (\S\ref{sec:exp_setup}) at each tool call, without (dotted) and with \textsc{LiveLedger} (solid). Numbers on the right are the shares at the end of the budget. With \textsc{LiveLedger}, fewer questions remain stalled at the end of the budget for all six agents.}
    \label{fig:ll_stalled}
    \vspace{-3mm}
\end{figure*}

\subsection{Effect of Explicit Verification States}\label{sec:exp_liveledger}

We expose the constraint states from \textsc{LiveLedger} (\S\ref{sec:liveledger}) to six agents (two trained and four prompt-based) and measure how their answers differ. 

\paragraph{Explicit states improve accuracy and verification across all six agents.}
Table~\ref{tab:liveledger-effect} compares Acc and UAR with and without \textsc{LiveLedger}. Exposing constraint states improves Acc by 4.4--16.1 points and reduces UAR by 6.2--16.7 points. These gains extend across both trained and prompt-based agents, showing that making verification progress explicit can improve answer accuracy and substantiation without retraining the search agents.

\paragraph{With constraint states exposed, agents verify more constraints and leave fewer unchecked.}
Figure~\ref{fig:ll_condition} compares the constraint outcomes with and without \textsc{LiveLedger}, split by the run result as in Figure~\ref{fig:exp_failure_modes}.
For five of the six agents, the share of verified constraints is 2.2--9.4 points higher and the share of unchecked constraints is 4.1--10.5 points lower, while the shares of assumed and refuted constraints shift by at most 2.4 points.
By contrast, larger or stronger models in Figure~\ref{fig:exp_failure_modes}(b, c) bring 4.8--6.5 points more assumed constraints with their higher Acc.
This suggests that knowing their own progress matters as much as a larger or stronger model.

For GLM-5.2, the exposed states appear to make the agent affirm fewer constraints it has not verified: its assumed and refuted constraints are 2.0 and 1.2 points lower, while the verified constraints are unchanged and the unchecked ones are 3.2 points higher.
The agent also appears to answer correctly on more of what it has verified: with the states exposed, its verified constraints in correct answers are 3.6 points higher (47.0 to 50.6), with 4.4 points higher Acc.

\begin{table}[t]
\centering
\footnotesize
\setlength{\tabcolsep}{1.5pt}
\renewcommand{\arraystretch}{1.2}
\begin{tabularx}{\columnwidth}{@{}Xcccc@{}}
\toprule
 & \multicolumn{2}{c}{Acc ($\uparrow$)} & \multicolumn{2}{c}{UAR ($\downarrow$)} \\
\cmidrule(lr){2-3}\cmidrule(lr){4-5}
Agent & w/o & w/ LL & w/o & w/ LL \\
\midrule
\multicolumn{5}{c}{\textit{Trained}} \\
\midrule
Agents-A1 \citep{bai2026scalinghorizonparametersreaching} & \Acc{33.5} & \Acc{41.5} & \Uar{65.7} & \Uar{49.0} \\
TongyiDR \citep{team2025tongyi} & \Acc{34.7} & \Acc{39.3} & \Uar{64.0} & \Uar{55.9} \\
\midrule
\multicolumn{5}{c}{\textit{Prompt-based}} \\
\midrule
gpt-oss-20b \citep{agarwal2025gpt} & \Acc{22.3} & \Acc{30.6} & \Uar{70.9} & \Uar{62.2} \\
gpt-oss-120b \citep{agarwal2025gpt} & \Acc{23.6} & \Acc{34.5} & \Uar{70.7} & \Uar{55.8} \\
DeepSeek-V4 \citep{deepseekai2026v4} & \Acc{43.6} & \Acc{59.7} & \Uar{55.6} & \Uar{41.5} \\
GLM-5.2 \citep{glm5team2026glm5vibecodingagentic} & \Acc{54.1} & \Acc{58.5} & \Uar{52.1} & \Uar{45.9} \\
\midrule
\multicolumn{5}{c}{\textit{Qwen2.5-7B: process rewards vs.\ LiveLedger$^\dagger$}} \\
\midrule
HiPRAG \citep{wu2025hiprag} & \Acc{18.1} & -- & \Uar{77.3} & -- \\
ReSeek \citep{li2025reseek} & \Acc{13.3} & -- & \Uar{81.8} & -- \\
Qwen2.5-7B & \Acc{16.6} & \Acc{21.0} & \Uar{80.2} & \Uar{74.6} \\
\bottomrule
\end{tabularx}
\caption{\textbf{Effect of \textsc{LiveLedger}.} Acc and UAR without (w/o) and with \textsc{LiveLedger} (w/ LL). Darker cells mean higher values. $^\dagger$HiPRAG and ReSeek train Qwen2.5-7B with process rewards.
}
\label{tab:liveledger-effect}
\end{table}

\paragraph{Fewer questions remain stalled.}
A run is stalled at a step if it has not yet answered and the candidate it finally commits to has gained no \textsc{Satisfied} constraint over the last three steps (\S\ref{sec:exp_setup}).

Figure~\ref{fig:ll_stalled} tracks the share of stalled questions at each tool call. Without the states, 37.8--57.6\% of questions are already stalled at the 10th tool call, and 20.5--36.0\% remain stalled at the end of the budget. When the constraint states are exposed, the share is lower for all six agents from the 20th tool call on (by 4.3--12.2 points) and 6.8--10.4 points lower at the end, and runs end 0.6--6.3 steps earlier on average. 
This suggests that knowing the constraint states keeps the agents making progress on their candidates, and that without them, the agents often lose track of which constraints remain open.

\paragraph{Explicit states outperform the tested process-reward baselines.}
The last group of Table~\ref{tab:liveledger-effect} compares \textsc{LiveLedger} with HiPRAG and ReSeek, which train Qwen2.5-7B with process rewards.
Without training the agent, Qwen2.5-7B with \textsc{LiveLedger} achieves 2.9 and 7.7 points higher Acc than HiPRAG and ReSeek, and 2.7 and 7.2 points lower UAR.
Acc in this group follows our earlier protocol (the mean of three runs under an earlier judge), so we compare it only within the group.
These results show that explicit constraint states can improve accuracy and substantiation without retraining the search agent.
\section{Conclusion}

In this work, we study \emph{illusory completion}, where agents conclude that a multi-constraint question is complete while a constraint of their answer remains unverified, even when the answer is correct.
This is critical, since a correct answer may rest on an unverified constraint, and the same habit produces wrong answers elsewhere.
Using the \textsc{Epistemic Ledger}, which tracks at every turn what the evidence establishes about each constraint and what the agent states about it, we find that unchecked constraints are the most common failure across 13 agents, and that training and scale shift constraints between failures rather than reducing them all.

To test whether making verification progress explicit can reduce these failures, we use \textsc{LiveLedger}, a lightweight 4B tracker, as a proxy for evidential states. Exposing these states to six agents improves accuracy by 4.4--16.1 points and increases fully verified answer rates by 6.2--16.7 points. These gains suggest that unreliable tracking or use of verification progress contributes to illusory completion.

Moving forward, we encourage benchmarks to report a diagnosis of progress, such as ours, alongside accuracy, which alone can mislead: the baseline with the highest accuracy assumes more constraints than any other prompt-based agent we test.
Training rewards, in turn, should credit an agent for knowing the state of each constraint, not only for answering correctly or reasoning consistently across steps.

\section*{Limitations} 
We acknowledge several limitations.
First, the \textsc{Epistemic Ledger} relies on an LLM evaluator, which can misjudge evidence or miss a belief the agent states; a missed affirmation, for instance, records an assumed constraint as unchecked. We validate the evaluator against human judgments and an independent evaluator (Appendix~\ref{sec:appx_validation}).
Also, \textsc{LiveLedger} approximates the constraint states, so the reported gap is an estimate.
Finally, we focus on questions whose constraints can be verified independently, leaving failures that arise from chained reasoning, as in multi-hop questions, to future work.


\bibliography{example_paper}


\clearpage
\appendix
\section{Experiment Details}\label{sec:appx_experiment_details}

\subsection{Dataset}\label{sec:appx_dataset}
We prompt \texttt{gpt-oss-120b} to decompose each question into a DAG whose nodes are the entities in the question. Each node lists the constraints that restrict the identity of its entity and the entities it depends on.
We keep a question if its DAG has a single entity node (depth $=1$) with three or more constraints (width $\geq 3$).
Table~\ref{tab:appx_dataset} lists the sources and the number of questions kept.
We keep the first 100 of the 185 BioASQ questions so that BioASQ does not outweigh the other benchmarks.
All 484 questions are in English.

\begin{table}[!ht]
\centering
\footnotesize
\setlength{\tabcolsep}{3pt}
\begin{tabularx}{\columnwidth}{@{}lXr@{}}
\toprule
Benchmark & Source split & \# \\
\midrule
BrowseComp & test & 205 \\
DeepSearchQA & eval, single-answer questions & 69 \\
FRAMES & test & 33 \\
LiveDRBench & v1-full, test & 49 \\
WebWalkerQA & main, English questions & 28 \\
BioASQ & \texttt{rag-mini-bioasq}, test (first 100) & 100 \\
\midrule
Total & & 484 \\
\bottomrule
\end{tabularx}
\caption{\textbf{Evaluation set.} Number of questions kept from each benchmark.}
\label{tab:appx_dataset}
\end{table}

\subsection{Computation}\label{sec:appx_computation}
We serve the open-weight agents with vLLM on NVIDIA GPUs.
In the shared runner, \texttt{gpt-oss-20b} uses one RTX A6000 (48GB), \texttt{gpt-oss-120b} uses two L40S (48GB) with tensor parallelism, and Agents-A1 uses one RTX PRO 6000 Blackwell (96GB) for each setting.
TongyiDR and WebExplorer run on L40S, RTX A6000, A40, and RTX PRO 6000 GPUs.
Search-o1 runs offline with vLLM on A100 (40GB) GPUs, two for \texttt{gpt-oss-20b} and four for \texttt{gpt-oss-120b}.
\textsc{LiveLedger} runs on one L40S, or on one RTX A6000 for DeepSeek-V4-Pro.
We access DeepSeek-V4-Pro and GLM-5.2 through OpenRouter, restricted to FP8 providers, and the evaluator, the judge, and the page-summary model through the OpenAI API.

\subsection{Validation}\label{sec:uar_human_evaluation}\label{sec:appx_validation}
We validate the \textsc{Epistemic Ledger}, with \texttt{gpt-5-nano} as the evaluator, against human judgments and an independent evaluator, and \textsc{LiveLedger} against this evaluator.   

\paragraph{Human evaluation.}
At the answer level, we sample 100 questions, evenly split between substantiated and unsubstantiated answers, restricted to runs with fewer than 10 tool calls due to the task length limits of Amazon Mechanical Turk.
Three annotators judge whether each answer satisfies every constraint given the agent's search results, and each is paid \$2 per task.
Individual annotations agree with the evaluator on 90\% of the questions and majority votes on 91\% (Fleiss' $\kappa=0.76$).

At the constraint level, on 100 human-annotated constraints, the evaluator's labels match the annotators on 88\% for evidential support $E$ and 93\% for stated belief $B$, and on a separate set of 100 cases, its outcomes match on 90\% for assumed, 84\% for refuted, 83\% for stalled, and 89\% for unchecked constraints.

\paragraph{Independent evaluator.}
Replacing the evaluator with \texttt{Qwen3.5-122B}, the two agree on 86.9\% of $E$ labels and 93.1\% of $B$ labels, and the failure rates derived from them differ by 0.0--2.9 points.

\paragraph{\textsc{LiveLedger}.}
To compare \textsc{LiveLedger} with \texttt{gpt-5-nano} on the same input, we take 120 trajectories of agents run without \textsc{LiveLedger} (60 each from \texttt{gpt-oss-120b} and DeepSeek-V4-Pro) and feed them to \textsc{LiveLedger} step by step, giving it the constraints that \texttt{gpt-5-nano} uses.   

For each constraint of the \textsc{Committed} candidate, the two assign the same final state to 82.5\% of the constraints (Gwet's AC1 $=0.80$), and 93.2\% of the constraints \textsc{LiveLedger} marks \textsc{Satisfied} are also \textsc{Satisfied} for \texttt{gpt-5-nano}.   

\subsection{Model Specification}\label{sec:appx_model_spec}
Table~\ref{tab:appx_models} lists the checkpoints we use.
Open-weight models are from HuggingFace, and DeepSeek-V4-Pro and GLM-5.2 are accessed on OpenRouter.
Agents-A1 is a mixture-of-experts model with 35B parameters, 3B of them active.
WebExplorer and TongyiDR summarize browsed pages with \texttt{gpt-4o-mini}.
\textsc{LiveLedger} is the 4B model described in Appendix~\ref{sec:appx_liveledger_details}.
The \textsc{Epistemic Ledger} evaluator is \texttt{gpt-5-nano}, and the judge for Acc is \texttt{gpt-5.6-sol}.

\begin{table*}[t!]
\centering
\footnotesize
\setlength{\tabcolsep}{4.5pt}
\renewcommand{\arraystretch}{1.1}
\begin{tabular}{@{}llllccr@{}}
\toprule
Agent & Cap per run & Search & Page reading & $T$ & top-$p$ & Max tokens \\
\midrule
Search-R1 & -- & Jina Search & none & 0.7 & -- & 1,024 \\
ASearcher & 30 model calls & Serper, top 10 & disabled & 0.6 & 0.95 & 4,098 \\
RAG-R1 & 10 retrievals & Jina Search & none & 0.7 & -- & 2,048 \\
DR-Tulu & -- & Serper, top 10 & Jina & 1.0 & -- & 32,000 \\
WebExplorer & 30 model calls & Serper, top 10 & Jina, summarized$^\ast$ & 0.6 & 0.95 & 10,000 \\
TongyiDR & 30 model calls & Serper, top 10; Scholar & Jina, summarized$^\ast$ & 0.85 & 0.95$^\dagger$ & 10,000 \\
Search-o1 & 30 turns & Jina Search, top 10 & 512-char windows & -- & -- & -- \\
\texttt{gpt-oss-20b} / \texttt{120b} & 30 agent calls & Serper, top 10 & Jina, 8,000 chars & 1.0 & 1.0 & 8,192 \\
DeepSeek-V4-Pro, GLM-5.2 & 30 agent calls & Serper, top 10 & Jina, 8,000 chars & 1.0 & -- & 8,192 \\
Agents-A1 & 30 agent calls & Serper, top 10 & Jina, 8,000 chars & 1.0$^\ddagger$ & 1.0$^\ddagger$ & 8,192 \\
\midrule
HiPRAG & 20 steps & Serper, top 10 & none & 0.7 & -- & 1,024 \\
ReSeek & 20 steps & Jina Search & none & 0.7 & -- & 2,048 \\
Qwen2.5-7B & 30 turns & Serper, top 10 & Jina, 8,000 chars & 1.0 & 1.0 & -- \\
\bottomrule
\end{tabular}
\caption{\textbf{Per-agent settings.} An agent call may issue several tool calls. ``--'' marks a setting that is not set or that we could not confirm from the run logs. $^\ast$By \texttt{gpt-4o-mini}. $^\dagger$With presence penalty 1.1. $^\ddagger$vLLM defaults; no sampling parameters are sent. The last group follows our earlier protocol.}
\label{tab:appx_agent_settings}
\end{table*}

\begin{table}[t!]
\centering
\footnotesize
\setlength{\tabcolsep}{3pt}
\begin{tabularx}{\columnwidth}{@{}l>{\raggedright\arraybackslash}X@{}}
\toprule
Agent & Checkpoint \\
\midrule
Search-R1 & \path{PeterJinGo/SearchR1-nq_hotpotqa_train-qwen2.5-7b-em-ppo} \\
ASearcher & \path{inclusionAI/ASearcher-Web-7B} \\
RAG-R1 & \path{yaoyueduzhen/RAG-R1-mq-7b} \\
DR-Tulu & \path{rl-research/DR-Tulu-8B} \\
WebExplorer & \path{hkust-nlp/WebExplorer-8B} \\
TongyiDR & \path{Alibaba-NLP/Tongyi-DeepResearch-30B-A3B} \\
Agents-A1 & \path{InternScience/Agents-A1} \\
\texttt{gpt-oss-20b}$^\ast$ & \path{openai/gpt-oss-20b} \\
\texttt{gpt-oss-120b}$^\ast$ & \path{openai/gpt-oss-120b} \\
DeepSeek-V4-Pro & \path{deepseek/deepseek-v4-pro} \\
GLM-5.2 & \path{z-ai/glm-5.2} \\
\midrule
HiPRAG & \path{qualidea1217/Qwen2.5-7B-Instruct-GRPO-HiPRAG} \\
ReSeek & \path{TencentBAC/ReSeek-qwen2.5-7b-em-grpo} \\
Qwen2.5-7B & \path{Qwen/Qwen2.5-7B-Instruct} \\
\bottomrule
\end{tabularx}
\caption{\textbf{Checkpoints.} $^\ast$Also the two Search-o1 backbones. The last group follows our earlier protocol.}
\label{tab:appx_models}
\end{table}

\subsection{Hyperparameters}\label{sec:appx_hyperparameters}

\paragraph{Agents that call tools directly.}
\texttt{gpt-oss-20b}, \texttt{gpt-oss-120b}, DeepSeek-V4-Pro, GLM-5.2, and Agents-A1 share one runner with two tools.
\texttt{search} takes a list of queries and returns the top 10 Serper results for each (title, URL, and snippet).
\texttt{browse} takes a list of URLs and returns each page from the Jina Reader, cut to 8,000 characters.
A run makes at most 30 agent calls that issue tool calls.
One call may issue several tool calls, which DeepSeek-V4-Pro (up to six) and GLM-5.2 (up to four) often do.
Each call generates up to 8,192 tokens with high reasoning effort, set through \texttt{reasoning\_effort} for \texttt{gpt-oss}, OpenRouter's \texttt{reasoning.effort} for DeepSeek-V4-Pro and GLM-5.2, and thinking mode for Agents-A1.
\texttt{gpt-oss}, DeepSeek-V4-Pro, and GLM-5.2 sample at temperature 1.0.
Agents-A1 receives no sampling parameters, so vLLM's defaults apply (temperature 1.0, top-$p$ 1.0).
The reasoning of earlier calls is passed back to Agents-A1 but not to the other models.
A failed agent call is retried up to five times.
For \texttt{gpt-oss}, the tools also accept common alternative argument names and return an error message for a malformed call.

\paragraph{\textsc{LiveLedger}.}
In this runner, \textsc{LiveLedger} samples at temperature 1.0 with up to 8,192 tokens and thinking off.
It extracts the constraints from the question at the start of a run.
After each agent call that uses a tool, it updates the ledger from the agent's reasoning (cut to 12,000 characters) and the latest tool result (cut to 40,000 characters), with at most 10 entries per update.
The ledger is shown to the agent as a table in a user message after the tool results and replaces the previous one.
Updates run in the background: before sending the update for call $t$, the runner waits for the update of call $t-1$ and shows it, so the agent's call $t+1$ sees the ledger updated through call $t-1$.

\paragraph{Other agents.}
For the other agents, we follow each agent's released inference code.
Table~\ref{tab:appx_agent_settings} lists the settings that differ across agents.
Search-o1 searches with Jina Search (top 10), fetches each result page, and inserts a 512-character window around the snippet without summarizing it.
Search-R1, RAG-R1, and ReSeek search with Jina Search and read only snippets.
ASearcher can request pages, but page access through Jina is disabled in its configuration, so it reads only Serper snippets.
TongyiDR can also search Google Scholar and run Python.
For the earlier-protocol group, Qwen2.5-7B uses the same \texttt{search} and \texttt{browse} tools as above, and Qwen2.5-7B with \textsc{LiveLedger} uses a \textsc{LiveLedger} checkpoint (step 52) from the fine-tuning run in Appendix~\ref{sec:appx_liveledger_details}, with thinking on and synchronous updates.

\paragraph{Evaluator and judge.}
The \textsc{Epistemic Ledger} evaluator, \texttt{gpt-5-nano}, uses medium reasoning effort, reads the first 30 steps of each trajectory, and retries a call up to five times when its output cannot be parsed.
The judge, \texttt{gpt-5.6-sol}, uses the default API settings and a separate prompt for BioASQ.

\FloatBarrier
\section{Results over Three Runs}\label{sec:appx_three_runs}
Table~\ref{tab:three-runs} reports Acc and UAR on the 484 questions for TongyiDR, \texttt{gpt-oss-20b}, and \texttt{gpt-oss-120b}, each run three times without and with \textsc{LiveLedger}, as the mean and standard deviation over the three runs.
For all three agents, every run with \textsc{LiveLedger} has higher Acc and lower UAR than every run without it.

\begin{table}[!ht]
\centering
\footnotesize
\setlength{\tabcolsep}{2pt}
\renewcommand{\arraystretch}{1.15}
\begin{tabularx}{\columnwidth}{@{}Xcccc@{}}
\toprule
 & \multicolumn{2}{c}{Acc ($\uparrow$)} & \multicolumn{2}{c}{UAR ($\downarrow$)} \\
\cmidrule(lr){2-3}\cmidrule(lr){4-5}
Agent & w/o & w/ LL & w/o & w/ LL \\
\midrule
TongyiDR      & 35.3{\scriptsize$\pm$0.5} & 42.1{\scriptsize$\pm$2.8} & 64.1{\scriptsize$\pm$0.5} & 52.2{\scriptsize$\pm$3.3} \\
gpt-oss-20b   & 23.4{\scriptsize$\pm$1.4} & 28.5{\scriptsize$\pm$2.0} & 70.6{\scriptsize$\pm$1.1} & 61.3{\scriptsize$\pm$1.1} \\
gpt-oss-120b  & 22.6{\scriptsize$\pm$1.0} & 34.2{\scriptsize$\pm$0.4} & 71.4{\scriptsize$\pm$1.0} & 56.4{\scriptsize$\pm$0.6} \\
\bottomrule
\end{tabularx}
\caption{\textbf{Results over three runs.} Acc (\%) and UAR (\%) on the 484 questions without (w/o) and with \textsc{LiveLedger} (w/ LL), as mean$\pm$standard deviation over three runs of each agent.}
\label{tab:three-runs}
\end{table}

\FloatBarrier
\section{Further Details of \textsc{LiveLedger}}\label{sec:appx_liveledger_details}

\subsection{Training Details}

\paragraph{Training Data Construction.} We fine-tune the \textsc{LiveLedger} based on Qwen3.5-4B. We (i) generate the SFT dataset using a teacher model (\texttt{gpt-oss-120b}), (ii) filter the datapoints with verification, and then (iii) distill the data to the student model. In the first stage, the teacher model is given questions from HDS-QA parallel questions and MultiConIR questions to perform inference. During inference, the ReAct search agent conducts multi-turn search reasoning while \textsc{LiveLedger} with \texttt{gpt-oss-120b} performs constraint extraction and constraint-state updates. Then, the generated constraint-extraction and constraint-state-update outputs are filtered: they are selected if the final answer is correct and deemed proper based on an additional verification task. In the additional verification task, \texttt{gpt-oss-120b} is prompted to judge whether the constraint extraction is correct based on the question, and whether the constraint-state update output is correct based on the search results and the updated ledger. To mitigate the long-tail distribution of single-entity Ledger-Update samples, we cap the entity-1 ratio at 0.20 by random subsampling. The resulting training dataset consists of 1,393 constraint extraction and 2,885 ledger update samples, totaling 4,278 instances.

\paragraph{Training Hyperparameters.} We split the dataset into train/val with a 0.9/0.1 ratio. We fine-tune Qwen3.5-4B (4.54B parameters, full fine-tuning) using ms-swift. We train with bf16 precision, the AdamW optimizer, learning rate 3e-5, a cosine schedule with 5\% warmup, weight decay 0.1, and gradient clipping at 1.0. The per-device train batch size is 1 with gradient accumulation 8, and we train on 16 GPUs, yielding an effective global batch size of 128. We train for 10 epochs (260 total steps) with a maximum sequence length of 8,192 tokens. Distributed training uses DeepSpeed ZeRO Stage 3 with CPU offloading for both optimizer states and parameters. The random seed is 42 for both data shuffling and training. We select the final model from the epoch-2 checkpoint based on validation loss.

\subsection{Implementation Details}

\textsc{LiveLedger} uses a separate 4B tracker $S$ to maintain constraint-level evidential states for both trained and prompt-based search agents. Given a question $q$, it extracts the constraints $\hat{C}$ and initializes the ledger $\hat{L}_0$. For each tool-using
step $t$, the tracker updates the ledger from the previous state and the new reasoning, action, and observation:

\[
\hat{L}_{t+1}
\sim S\!\left(
\cdot \mid q, \hat{C}, \hat{L}_t, \tau_t, a_t, o_t
\right).
\]

The tracker grounds constraint states in retrieved evidence rather than treating the agent's assertions as evidence.

The search agent $\pi$ does not generate ledger updates. Let $\hat{L}^{\mathrm{vis}}_t$ denote the ledger supplied to the agent after step $t$. The agent then samples its next reasoning and action:

\[
\tau_{t+1}, a_{t+1}
\sim \pi\!\left(
\cdot \mid H_t, \hat{L}^{\mathrm{vis}}_t
\right).
\]

In the shared runner, the ledger is presented as a table in a user message after the tool results, replacing the previous table. Updates run asynchronously, so the ledger supplied to the agent may lag behind the latest observation. Update timing and per-agent settings are detailed in Appendix~\ref{sec:appx_hyperparameters}.

\begin{figure}[t!]
    \centering
    \includegraphics[width=\linewidth]{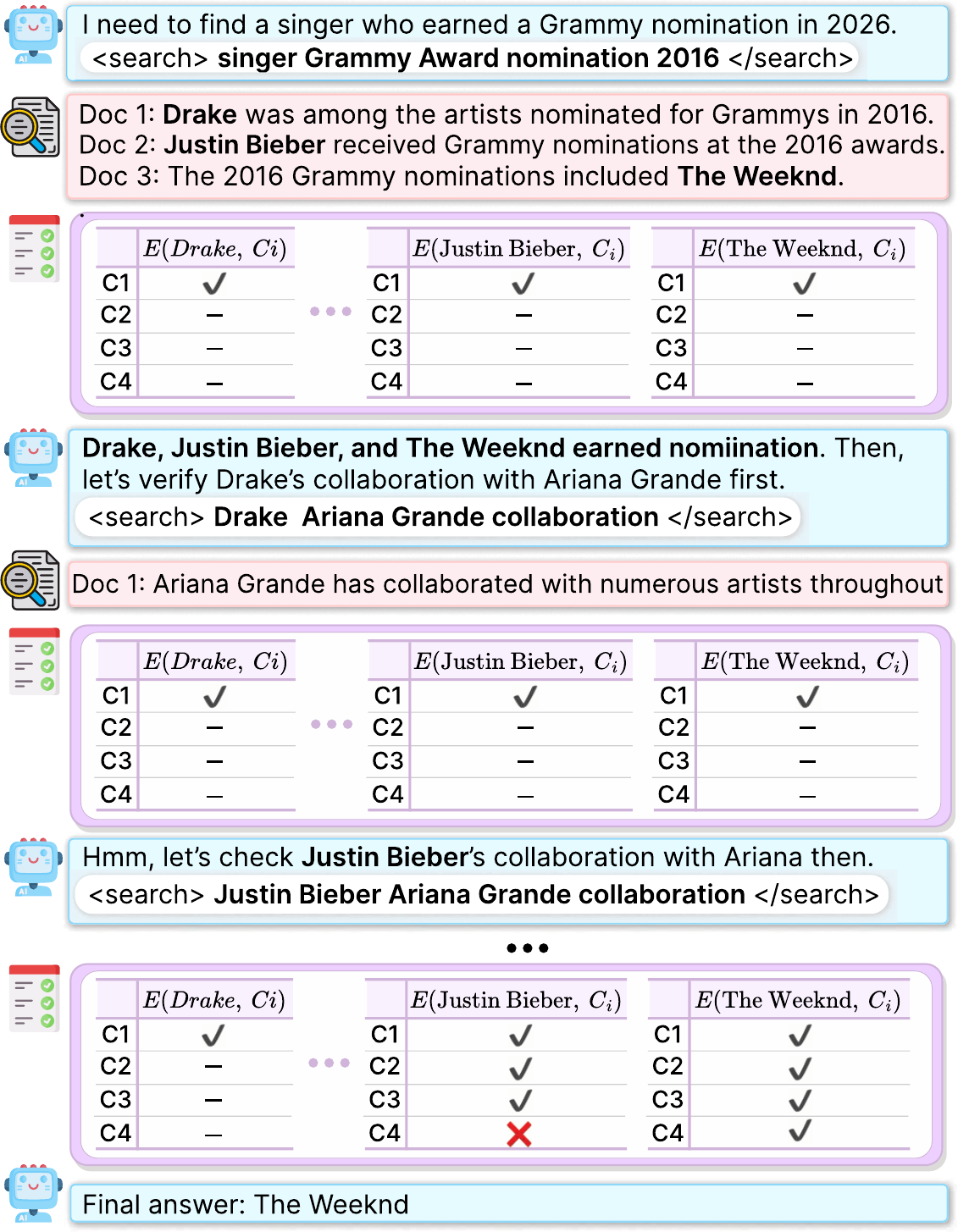} 
    \caption{An illustration of \textsc{LiveLedger} integration into the agent.}
    \label{fig:method-appx}
\end{figure}

\FloatBarrier
\section{Cost}\label{sec:appx_cost}
\paragraph{\textsc{Epistemic Ledger}.}
Table~\ref{tab:appx_cost} reports the cost of the evaluator per question.
\texttt{gpt-5-nano} makes 29--57 calls per question, reading 131--223K input tokens and writing 148--249K output tokens, most of which (84--89\%) are reasoning tokens.
This costs \$0.065--0.108 per question at list prices, or \$31--52 for one agent on all 484 questions.
The evaluator reads 0.34--0.61 times as many input tokens as the agent itself.

\paragraph{\textsc{LiveLedger}.}
The 4B tracker is called 12--21 times per question, reading 69--127K tokens and writing 2.3--3.9K tokens in total.
Only constraint extraction and the wait for pending updates block the agent; the per-step updates run while the agent continues.
The agent waits on the tracker for 24--97 seconds per question, 1.1--8.4\% of its end-to-end latency (Table~\ref{tab:appx_cost}).

\begin{table}[!ht]
\centering
\footnotesize
\setlength{\tabcolsep}{2.5pt}
\begin{tabularx}{\columnwidth}{@{}Xcccc@{}}
\toprule
 & \multicolumn{2}{c}{\textsc{Epistemic Ledger}} & \multicolumn{2}{c}{\textsc{LiveLedger}} \\
\cmidrule(lr){2-3}\cmidrule(lr){4-5}
Agent & Calls & USD & Calls & Wait (\%) \\
\midrule
TongyiDR      & 48.2 & 0.093 & 20.9 & 7.4 \\
Agents-A1     & 50.5 & 0.108 & 19.4 & 5.5 \\
gpt-oss-20b   & 52.3 & 0.089 & 19.7 & 2.0 \\
gpt-oss-120b  & 57.2 & 0.095 & 18.7 & 1.1 \\
DeepSeek-V4-Pro & 46.8 & 0.089 & 15.7 & 8.4 \\
GLM-5.2       & 32.6 & 0.065 & 12.0 & 4.1 \\
\bottomrule
\end{tabularx}
\caption{\textbf{Cost per question.} \textsc{Epistemic Ledger}: evaluator calls and cost at \texttt{gpt-5-nano} list prices, for the agent without \textsc{LiveLedger}. \textsc{LiveLedger}: tracker calls and the share of end-to-end latency the agent spends waiting on the tracker.}
\label{tab:appx_cost}
\end{table}

\FloatBarrier
\section{Qualitative Examples}\label{sec:appx_qualitative_examples}

\begin{figure*}[p]
\vspace{-2mm}
\caption{Example of an \textbf{\textcolor{red}{incorrect}} agent execution, abbreviating the agent's reasoning text for space.}
\vspace{-3mm}
\centering
\begin{tcolorbox}[colback=white, colframe=black!30, boxrule=2pt, arc=2pt]

\noindent
\footnotesize
\textbf{Question:} As of August 3, 2024, which rabbi worked for both Reform Congregation Keneseth Israel in Philadelphia and Congregation Beth Israel in West Hartford, Connecticut?

\vspace{0.35cm}
\noindent
\textbf{Gold:} Abraham J. Feldman \quad \textbf{Prediction:} Rabbi Stephen Lewis Fuchs \  {\XSolidBrush} 

\vspace{0.35cm}
\hrule
\vspace{0.35cm}

\noindent
\begin{tabular}{@{}p{1.5cm}!{\color{white}\vrule width 1.5pt}p{\dimexpr\textwidth-2.6cm\relax}@{}}

\footnotesize
\textbf{\textsc{\textcolor{teal!90}{Assistant}}}  & 
\footnotesize
Query: [``Congregation Keneseth Israel Philadelphia rabbi list``, ``Congregation Keneseth Israel Philadelphia Rabbi``, ``Keneseth Israel Philadelphia Rabbi history``, ``Congregation Beth Israel West Hartford rabbi history``, ``Congregation Beth Israel West Hartford CT rabbi``] 
\vspace{0.35cm}
\\

\footnotesize
\textbf{\textsc{Tool}} & 
\footnotesize
\textbf{Query 1-3:} [1] Rabbi Amy Levy Cantor, Lance J. Sussman Ph.D Rabbi Emeritus ... [2] Since appointing its first rabbi in 1861, the congregation has been led by eight rabbis, including David Einhorn, Joseph Krauskopf, Bertram Korn, and Simeon ... [3] In 1924, Dr. William H. Fineshriber was called to the pulpit of Keneseth Israel ... Joseph Krauskopf (1858-1923), was the third senior rabbi ...
\textbf{Query 4-5:} [1] We moved to our present home in West Hartford in 1936 ... [2] {\setlength{\fboxsep}{1pt}\colorbox{yellow!20}{Rabbi Stephen Fuchs.} Rabbi Fuchs served as Senior Rabbi of \colorbox{orange!10}{Congregation Beth Israel from 1997-2011}. Following his retirement from Beth Israel ...} [3] {\setlength{\fboxsep}{1pt}\colorbox{yellow!20}{Rabbi Stephen Fuchs} (Emeritus)} ... 
\vspace{0.35cm}
\\

\footnotesize
\textbf{\textsc{\textcolor{teal!90}{Assistant}}} & 
\footnotesize
Query: [``Stephen Fuchs`` ``Keneseth Israel``, "Stephen Fuchs`` ``Reform Congregation Keneseth Israel``] 
\vspace{0.35cm}
\\

\footnotesize
\textbf{\textsc{Tool}} & 
\footnotesize
\textbf{Query 1-2:} [1] It was commissioned and presented by {\setlength{\fboxsep}{1pt}\colorbox{cyan!10}{Reform Congregation Keneseth Israel} (KI) in Elkins Park, Philadelphia ... \colorbox{red!10}{Rabbi Stephen Fuchs' sermon} on.} [2] {\setlength{\fboxsep}{1pt}\colorbox{yellow!20}{Rabbi Stephen Fuchs,} Bat Yam Temple of the Islands, Sanibel ... Rabbi Stacy Rigler, \colorbox{cyan!10}{Reform Congregation Keneseth Israel,} Elkins Park.} [3] {\setlength{\fboxsep}{1pt}\colorbox{yellow!20}{Rabbi Stephen Fuchs} officiated at the wedding ... \colorbox{cyan!10}{Keneseth Israel, Philadelphia, Pennsylvania} ...}
\vspace{0.35cm}
\\

\footnotesize
\textbf{\textsc{\textcolor{teal!90}{Assistant}}} & 
\footnotesize
Query: [``Stephen Fuchs`` ``Keneseth Israel`` Philadelphia``, ``Stephen Fuchs`` ``Reform Congregation Keneseth Israel`` Philadelphia``] 
\vspace{0.35cm}
\\

\footnotesize
\textbf{\textsc{Tool}} & 
\footnotesize
\textbf{Query 1-2:} [1] It was commissioned and presented by {\setlength{\fboxsep}{1pt}\colorbox{cyan!10}{Reform Congregation Keneseth Israel} (KI) in Elkins Park, Philadelphia ... \colorbox{red!10}{Rabbi Stephen Fuchs' sermon} on.} [2] {\setlength{\fboxsep}{1pt}\colorbox{yellow!20}{Rabbi Stephen Fuchs,} Bat Yam Temple of the Islands, Sanibel ... Rabbi Stacy Rigler, \colorbox{cyan!10}{Reform Congregation Keneseth Israel,} Elkins Park.} [3] {\setlength{\fboxsep}{1pt}\colorbox{yellow!20}{Rabbi Stephen Fuchs} officiated at the wedding ... \colorbox{cyan!10}{Reform Congregation Keneseth Israel, Philadelphia, Pennsylvania} ...}
\vspace{0.35cm}
\\

\footnotesize
\textbf{\textsc{\textcolor{teal!90}{Assistant}}} & 
\footnotesize
Query: [``Stephen Fuchs Keneseth``, ``Rabbi Stephen Fuchs Keneseth``] 
\vspace{0.2cm}
\\

 & \quad \quad \quad \quad \quad \quad \quad \quad \quad \quad \quad  \footnotesize\textbf{(Continued ...)} 
\vspace{0.2cm} 
\\

\footnotesize
\textbf{\textsc{\textcolor{teal!90}{Assistant}}} & 
\footnotesize
$\boxed{\text{Rabbi Stephen Lewis Fuchs}}$ \quad {\color{red}\textbf{INCORRECT}} \\
\end{tabular}
\vspace{-2mm}
\end{tcolorbox}
\label{fig:search_example_incorrect}
\end{figure*}




\begin{table*}[p]
\centering
\vspace{-1cm}
\caption{The \textsc{Epistemic Ledger} after the agent execution in Figure \ref{fig:search_example_incorrect}.}
\small
\resizebox{\textwidth}{!}{%
\begin{tabular}{lcc}
\textbf{Candidate}: Stephen Lewis Fuchs \\
\textbf{Status}: \textsc{\textbf{Active}} \vspace{1mm} \\
\hline
\textbf{Constraint} & \textbf{Evidential Support} & \textbf{Agent Belief} \\
\hline
C1: Individual must be a rabbi & \checkmark & \checkmark \\
 & \multicolumn{1}{p{5cm}}{\setlength{\fboxsep}{1pt}\colorbox{yellow!20}{Rabbi Stephen Lewis Fuchs}} & \multicolumn{1}{p{5cm}}{\setlength{\fboxsep}{1pt}\colorbox{yellow!20}{is Rabbi Stephen Lewis Fuchs}} \\
\hline
C2: Worked for Keneseth Israel & \checkmark & \checkmark \\
in Philadelphia & \multicolumn{1}{p{5cm}}{\setlength{\fboxsep}{1pt}\colorbox{cyan!10}{Rabbi Stephen Fuchs officiated at}} & \multicolumn{1}{p{5cm}}{\setlength{\fboxsep}{1pt}\colorbox{cyan!10}{served Reform Congregation}} \\
 & \multicolumn{1}{p{5cm}}{\setlength{\fboxsep}{1pt}\colorbox{cyan!10}{the wedding ... Reform Congregation}} & \multicolumn{1}{p{5cm}}{\setlength{\fboxsep}{1pt}\colorbox{cyan!10}{Keneseth Israel in Philadelphia}} \\
 & \multicolumn{1}{p{5cm}}{\setlength{\fboxsep}{1pt}\colorbox{cyan!10}{Keneseth Israel, Philadelphia, PA}} & \\
\hline
C3: Worked for Beth Israel & \checkmark & \checkmark \\
in West Hartford, CT & \multicolumn{1}{p{5cm}}{\setlength{\fboxsep}{1pt}\colorbox{orange!10}{served as Senior Rabbi at}} & \multicolumn{1}{p{5cm}}{\setlength{\fboxsep}{1pt}\colorbox{orange!10}{served Congregation Beth Israel}} \\
 & \multicolumn{1}{p{5cm}}{\setlength{\fboxsep}{1pt}\colorbox{orange!10}{Congregation Beth Israel, until 2011}} & \multicolumn{1}{p{5cm}}{\setlength{\fboxsep}{1pt}\colorbox{orange!10}{in West Hartford}} \\
\hline
C4: Both positions by & \XSolidBrush & --- \\
August 3, 2024 & \multicolumn{1}{p{5cm}}{\setlength{\fboxsep}{1pt}\colorbox{red!10}{In summer 2024 Rabbi Fuchs}} & \\
 & \multicolumn{1}{p{5cm}}{\setlength{\fboxsep}{1pt}\colorbox{red!10}{assumed position at Temple Beth}} & \\
 & \multicolumn{1}{p{5cm}}{\setlength{\fboxsep}{1pt}\colorbox{red!10}{Shalom, Vero Beach, FL}} & \\
\hline
\end{tabular}%
}
\end{table*}




\begin{figure*}[p]
\caption{Example of an \textbf{\textcolor{blue}{correct}} agent execution with \textbf{\textsc{LiveLedger}}, abbreviating the agent's reasoning text for space.}
\centering
\small
\begin{tcolorbox}[colback=white, colframe=black!30, boxrule=2pt, arc=2pt]

\noindent
\textbf{Question:} As of August 3, 2024, which rabbi worked for both Reform Congregation Keneseth Israel in Philadelphia and Congregation Beth Israel in West Hartford, Connecticut?

\vspace{0.35cm}
\noindent
\textbf{Gold:} Abraham J. Feldman \quad \textbf{Prediction:} Rabbi Abraham J. Feldman {\checkmark} 

\vspace{0.35cm}
\hrule
\vspace{0.35cm}

\noindent
\begin{tabular}{@{}p{1.5cm}!{\color{white}\vrule width 1.5pt}p{\dimexpr\textwidth-2.6cm\relax}@{}}

\textbf{\textsc{\textcolor{teal!90}{Assistant}}}  & Query: [``Reform Congregation Keneseth Israel Philadelphia rabbi``, ``Congregation Keneseth Israel Philadelphia rabbi history``, ``Congregation Beth Israel West Hartford rabbi``, ``Rabbi Congregation Beth Israel West Hartford Connecticut``, ``Rabbi Keneseth Israel Philadelphia``, ``Rabbi Beth Israel West Hartford``] 
\vspace{0.35cm}
\\

\textbf{\textsc{Tool}} & \textbf{Query 1-2:} [1] Our Clergy - Rabbi Amy Levy Cantor, Lance J. Sussman Ph.D Rabbi Emeritus ... Since appointing its first rabbi in 1861, the congregation has been led by eight rabbis, including David Einhorn, Joseph Krauskopf, Bertram Korn ... Dr. William H. Fineshriber was called to the pulpit ... 
\textbf{Query 3-4:} [1] Clergy - Rabbi Michael Pincus ... Rabbi Fuchs served as Senior Rabbi of Congregation Beth Israel from 1997-2011 ... 
\textbf{Query 5-6:} [1] Rabbi Benjamin P. David ... The current senior rabbi is Lance J. Sussman ... Rabbi Stephen Fuchs ... 
\vspace{0.35cm}
\\

\textbf{\textsc{\textcolor{teal!90}{Assistant}}} & Query: [``Rabbi ``Keneseth Israel`` ``Beth Israel`` ``West Hartford``] 
\vspace{0.35cm}
\\

\textbf{\textsc{Tool}} & \textbf{Query 1:} [1] {\setlength{\fboxsep}{1pt}\colorbox{yellow!20}{Rabbi Abraham J Feldman} - ... Rabbi at \colorbox{cyan!10}{Keneseth Israel}. Abraham Jehiel ... After serving at KI he became a \colorbox{yellow!20}{Rabbi} at \colorbox{orange!10}{Congregation Beth Israel in West Hartford, Connecticut (1925-1968)}. 
[2] Abraham J. Feldman - ... \colorbox{cyan!10}{Keneseth Israel in Philadelphia, Pennsylvania, from 1920 to 1925} ... He was elected rabbi of \colorbox{orange!10}{Congregation Beth Israel in West Hartford} ... 
[3] MS-38: Abraham J. Feldman Papers ... Keneseth Israel in ... \colorbox{orange!10}{Beth Israel in West Hartford, Connecticut (1925-1968)}. In 1968, he was elected Rabbi Emeritus ...}
\vspace{0.35cm}
\\

\textbf{\textsc{\textcolor{purple}{User}}} & \textbf{Ledger Update - Abraham J. Feldman:}
\vspace{1mm}
\newline

{\setlength{\tabcolsep}{2pt} 
\begin{tabularx}{\linewidth}{Xcl}
\hline
\textbf{Constraint} & \textbf{Status} & \textbf{Evidence} \\
\hline
C1: Individual must be a rabbi & \checkmark & {\setlength{\fboxsep}{1pt}\colorbox{yellow!20}{Rabbi Abraham J Feldman}} \\
C2: Worked for Keneseth Israel in Philadelphia & \checkmark & {\setlength{\fboxsep}{1pt}\colorbox{cyan!10}{1920-1925}} \\
C3: Worked for Beth Israel in West Hartford, CT & \checkmark & {\setlength{\fboxsep}{1pt}\colorbox{orange!10}{1925-1968}} \\
C4: Both positions by August 3, 2024 & \checkmark & {\setlength{\fboxsep}{1pt}\colorbox{orange!10}{Historical record confirmed}} \\
\hline
\end{tabularx}
}

\vspace{0.35cm}
\\

\textbf{\textsc{\textcolor{teal!90}{Assistant}}} & Browse: [``https://en.wikipedia.org/wiki/Abraham\_J.\_Feldman``] 
\vspace{0.35cm}
\\

\textbf{\textsc{Tool}} & \textbf{Abraham Jehiel Feldman} (June 28, 1893 -- July 21, 1977) was a Ukrainian-born American rabbi ... He then ministered at ... the Reform Congregation Keneseth Israel in Philadelphia, Pennsylvania, from 1920 to 1925. While he went to the latter congregation as an assistant rabbi under Rabbi Joseph Krauskopf ... He was elected rabbi of Congregation Beth Israel in West Hartford, Connecticut, in 1925, and he served as rabbi there until his retirement in 1968 ... 
\vspace{0.35cm}
\\

\textbf{\textsc{\textcolor{teal!90}{Assistant}}} & $\boxed{\text{Rabbi Abraham J. Feldman}}$\quad {\color{blue}\textbf{CORRECT}}  \\
\end{tabular}

\end{tcolorbox}
\label{fig:search_example}
\end{figure*}

\FloatBarrier
\section{Prompts}\label{sec:appx_prompt}

\paragraph{Prompt for Question Selection.}
\normalsize
This prompt instructs the model to decompose a given complex question into a structured representation in the form of a directed acyclic graph (DAG), by identifying entities, their defining constraints, and dependency relationships. Based on this structured output, we select questions whose DAG contains a single entity with no dependencies and more than two associated constraints.
\small
\begin{tcolorbox}[breakable, colframe=black, title={Prompt for Question Collection by 
Decomposing Questions and Extracting DAG Entity \& Dependency}]

\#\#\# Task Instruction:\\
You will be given a natural-language question. Your task is to analyze it by constructing a **Directed Acyclic Graph (DAG)** of entities and relations, following data-flow reasoning principles similar to those used in Beyond ReAct and other computation-graph approaches.\\

\#\#\# Goals:\\
1. Identify all **entities** mentioned or implied in the question.\\
2. Build a DAG where each entity is a **node**, and directed edges represent **dependency relations** (`depends\_on`).\\
3. For each entity, extract all explicit **constraints** that restrict the possible identity of that entity.\\
4. Output a JSON object containing each entity with a list of its constraints and a `depends\_on` list indicating which other entities determine it.\\

You **do not** need to solve the question.\\
You **only** need to produce the structured representation of the entities, dependencies, and constraints.\\

\#\#\# Definitions:\\

**1. Entity (Node)**\\
A distinct entity (person, object, organization, location, event, etc.) referenced in the question. Entities must be explicit and meaningfully participate in the question's logic.\\

**2. Dependency (Edge)**\\
A directed relation where one entity's identity **depends on** another entity.\\
Dependency direction: If X is determined by Y, then **X depends\_on Y**.\\
Ignore trivial dependencies like "full name of the actor" → "actor".\\

**3. Constraint**\\
A condition that reduces the set of possible candidates for an entity.\\
Assign constraints to the entity they restrict, not to other related nodes.\\
Not constraints include pure structural relations ("capital of…", "husband of…", "author of…") or non-restrictive statements.\\

\#\#\# Output Format (JSON only):\\
You must output a JSON object with this exact shape:\\

\{ "entities": \{ "<entity\_name>": \{ "constraints": ["...", "..."], "depends\_on": ["<other\_entity>"] \}, "<entity\_name\_2>": \{ "constraints": [...], "depends\_on": [...] \} \} \}\\

Guidelines:\\
- Use simple, descriptive keys for entity names (e.g., "building", "book", "author", "rank", "country", "president").\\
- If an entity has no constraints, return an empty list.\\
- If an entity depends on no other entities, return an empty list.\\
- Do not include fields other than `entities`, `constraints`, and `depends\_on`.\\
- Do not output explanatory text—only the JSON.\\
- **Minimize the number of entities** by merging simple lookups into constraints.\\
- **Preserve lookup chains** where each step requires distinct identification. Merge only trivial, well-known references into constraints.\\

\#\#\# Examples:\\

**Example 1**\\
**Question:** Who was born on September 3, 1937, is an alumnus of National Taiwan Normal University, and placed first in Group B of the Joint University Entrance Exam?\\
**Output:**\\
\{ "entities": \{ "person": \{ "constraints": ["Born on September 3, 1937", "Alumnus of National Taiwan Normal University", "Placed first in Group B of the Joint University Entrance Exam"], "depends\_on": [] \} \} \}\\

**Example 2**\\
**Question:** Name the monument that as of December 2023 satisfies a list of historical and spatial conditions.\\
**Output:**\\
\{ "entities": \{ "monument": \{ "constraints": ["Erected in the 19th Century", "Restored in the 21st Century", "Represents a political dynasty", "Has a name on it of a World War One soldier", "Hosted a food festival on the square it was built", "Once had an ironworks in the area", "Is in a town founded in the 900s", "Is 30 to 60 meters from a train station", "Is 30 to 60 meters from a Chinese restaurant", "Is 30 to 60 meters from a Department store", "Is 60 to 80 meters from a driver's license service"], "depends\_on": [] \} \} \}\\

**Example 3**\\
**Question:** Imagine there is a building called Bronte Tower whose height equals the Dewey Decimal classification of a Charlotte Brontë book published in 1847. Where would this building rank among tallest buildings in New York City as of August 2024?\\
**Output:**\\
\{ "entities": \{ "building": \{ "constraints": ["Building is named Bronte Tower", "Height equals the Dewey Decimal classification number of the referenced book"], "depends\_on": ["book"] \}, "book": \{ "constraints": ["Written by Charlotte Brontë", "Published in 1847"], "depends\_on": [] \}, "rank": \{ "constraints": ["Rank among tallest buildings in New York City as of August 2024"], "depends\_on": ["building"] \} \} \}\\

\#\#\# Final Instruction:\\
Analyze the following question by constructing a DAG of entities, assigning constraints to each entity, and specifying dependency relations.\\
**Remember:** Preserve lookup chains where each step requires distinct identification. Merge only trivial, well-known references into constraints.\\
Output **only** the JSON object described above.\\

\#\#\# Inputs:\\
- **Question:** \{question\}\\

\#\#\# Output:\\
Only the JSON object.\\
\end{tcolorbox}


\paragraph{Prompt for Constraint Extraction.}
\normalsize
This prompt directs the model to identify explicit, externally verifiable constraints stated in a question that any valid answer must satisfy. The model decomposes the question into atomic, non-overlapping conditions while excluding implicit assumptions and formatting requirements for initializing the epistemic ledger.
\begin{tcolorbox}[breakable, colframe=black, title={Prompt for Constraint Extraction}]
\#\#\# Task Instruction:\\
Please extract explicit, externally verifiable constraints from a question that an answer must satisfy.\\

Your task is to read a question and produce a constraint list in JSON format. Each constraint must describe a condition that the answer must satisfy and must be objectively verifiable using an external source (e.g., factual records or publicly available information).\\

\#\#\# Instructions:\\
1. Decompose the question into a list of atomic constraints that cannot be further decomposed into smaller conditions.\\
2. For each constraint:\\
\quad - It must be **explicitly mentioned** in the question.\\
\quad - It must be objectively verifiable using an external source (e.g., factual records or publicly available information).\\
\quad - **Exclude** implicit, assumed, or inferred constraints, as well as constraints about the **answer format** (e.g., length, structure, naming style).\\
3. Output **only** valid JSON in the specified format.\\

\#\#\# Output Format (JSON only):\\
\{ "constraint\_1": \{ "constraint": "<externally verifiable condition the answer must satisfy>" \}, "constraint\_2": \{ "constraint": "<externally verifiable condition the answer must satisfy>" \} \}\\

\#\#\# Examples:\\

**Example 1**\\
**Question:** Name a publicly traded technology company that was founded before 1990, is headquartered in Japan, and is listed on the Tokyo Stock Exchange.\\
**Output:**\\
\{ "constraint\_1": \{ "constraint": "The company must be a publicly traded technology company" \}, "constraint\_2": \{ "constraint": "The company must have been founded before 1990" \}, "constraint\_3": \{ "constraint": "The company must be headquartered in Japan" \}, "constraint\_4": \{ "constraint": "The company must be listed on the Tokyo Stock Exchange" \} \}\\

**Example 2**\\
**Question:** Name a film that won the Academy Award for Best Picture, was directed by a woman, and was released in the 2010s.\\
**Output:**\\
\{ "constraint\_1": \{ "constraint": "The film must have won the Academy Award for Best Picture" \}, "constraint\_2": \{ "constraint": "The film must have been directed by a woman" \}, "constraint\_3": \{ "constraint": "The film must have been released in the 2010s" \} \}\\

**Example 3**\\
**Question:** Name an Olympic athlete who has won gold medals in at least three different Summer Olympic Games and competed for the United States.\\
**Output:**\\
\{ "constraint\_1": \{ "constraint": "The person must be an Olympic athlete" \}, "constraint\_2": \{ "constraint": "The person must have won gold medals in at least three different Summer Olympic Games" \}, "constraint\_3": \{ "constraint": "The person must have competed representing the United States" \} \}\\

\#\#\# Inputs:\\
- **Question:** \{question\}\\

\#\#\# Output:\\
Only the JSON object.\\
\end{tcolorbox}

\paragraph{Prompt for \textsc{Epistemic Ledger} Updates.}
\normalsize
The two prompts below govern the structured updating of an epistemic ledger by separating objective evidence from the agent’s subjective beliefs during multi-step reasoning. It requires the model to independently record what is verified by external sources (\textit{obj}) and what is assumed or believed in the agent’s reasoning trace (\textit{per}), while tracking candidate status transitions across steps. This design enables fine-grained analysis of reasoning dynamics, including belief formation, revision, and divergence from verified evidence.
\begin{tcolorbox}[breakable, colframe=black, title={Prompt for Objective Evidence Ledger Update}]
\#\#\# Task Instruction:\\
You are an Objective Evidence Ledger Annotator. Your ONLY job is to update `null` values of `obj` and `obj\_evidence` for each Candidate × constraint using ONLY the Search Results.\\

\#\#\# Core Principle:\\
- `obj` must reflect what has been objectively found in the Search Results, not what the agent thinks.\\

\#\#\# Definitions:\\

**Candidate**\\
- Any entity proposed as a possible answer in the Search Results or reasoning trajectory must be treated as a Candidate.\\
- If multiple names refer to the same real-world entity, treat them as a single Candidate and aggregate evidence.\\
- Include all possible candidates uniquely in the ledger; do not miss or over-aggregate.\\
- Do not assume phrases in the question or constraints as candidates.\\

**obj (Objective)**\\
- `true` IF:\\
\quad - Search Results explicitly and unambiguously prove the exact claim made by the constraint; OR\\
\quad - The claim can be clearly inferred from Search Results without ambiguity and matches scope, subject, and meaning.\\
- `false` IF:\\
\quad - Contradiction: Search Results explicitly deny the claim; OR\\
\quad - Scope mismatch: evidence refers to a related but different fact and does NOT satisfy the constraint; OR\\
\quad - Exhaustion (optional): a clearly targeted query produces no proof (see Exhaustion Rule).\\
- `null` IF:\\
\quad - No relevant Search Results evidence has been observed.\\

**Evidence**\\
- If `obj` is `true` or `false`, `obj\_evidence` MUST be a verbatim snippet from the Search Results.\\
- If `obj` is `null`, `obj\_evidence` MUST be `null`.\\

\#\#\# Exhaustion Rule:\\
Mark `obj=false` by exhaustion ONLY when both conditions hold:\\
1. The Search Query is clearly targeted at verifying the constraint.\\
2. The Search Results contain no proof for that constraint for the Candidate.\\
If no verbatim snippet can justify `false`, keep `obj=null`.\\

\#\#\# Rules:\\
1. Update ONLY `obj` and `obj\_evidence`, using ONLY verbatim information from Search Results.\\
\quad - Update `null` values when evidence supports `true` or `false`.\\
\quad - Update existing `true` or `false` values ONLY if contradicted by new evidence.\\
\quad - If no relevant evidence exists, leave values unchanged.\\
2. Do NOT update `per`, `per\_evidence`, or `status`.\\
3. Candidates start with all constraints having `obj=null` unless proven otherwise.\\
4. Aggregate duplicate mentions of the same Candidate.\\
5. Assess each constraint independently.\\

\#\#\# Output Format (JSON only):\\
\{ "\verb=<=Candidate Name\verb=>=": \{ "status": "active$|$stored$|$rejected", "constraints": \{ "\verb=<=constraint\_id\verb=>=": \{ "obj": true$|$false$|$null, "per": null, "obj\_evidence": "\verb=<=verbatim snippet or null\verb=>=", "per\_evidence": null \} \} \} \}\\

\#\#\# Inputs:\\
- **Question:** \{question\}\\
- **Constraints:** \{constraints\}\\
- **Current Perception \& Status Ledger:** \{current\_ledger\}\\
- **Previous Thinking:** \{prev\_thinking\}\\
- **Search Query:** \{query\}\\
- **Search Results:** \{result\}\\
- **Next Thinking:** \{next\_thinking\}\\

\#\#\# Output:\\
Only the updated JSON ledger.\\
\end{tcolorbox}

\begin{tcolorbox}[breakable, colframe=black, title={Prompt for Perception \& Status Ledger Update}]
\#\#\# Task Instruction:\\
You are a Perception \& Status Ledger Annotator. Your ONLY job is to update candidate `status` and each constraint’s `per` and `per\_evidence`.\\
You MUST capture what the agent **believes** in the thinking text, even if it is wrong or unsupported.\\

\#\#\# Core Principle:\\
- `per` must be derived ONLY from the agent’s thinking (Previous Thinking + Next Thinking).\\
- Do NOT use Search Results content as evidence for `per`; use them only to notice candidate names.\\

\#\#\# Definitions:\\

**Candidate**\\
- Any entity proposed as a possible answer in Next Thinking must be treated as a Candidate.\\
- If multiple names refer to the same real-world entity, treat them as a single Candidate and aggregate.\\
- Include all possible candidates uniquely; do not miss or over-aggregate.\\
- Do not assume phrases in the question or constraints as candidates.\\

**Status**\\
- `active`: the agent is currently focusing on or selecting this candidate.\\
- `stored`: mentioned but not currently selected; still considered possible.\\
- `rejected`: the agent explicitly rules it out or abandons it as failing.\\

**per (Perceived)**\\
- `true`: the agent’s thinking assumes the constraint is satisfied.\\
- `false`: the agent’s thinking disqualifies the candidate or identifies failure on the constraint.\\
- `null`: the agent has not expressed a belief about the constraint.\\

**Evidence**\\
- If `per` is `true` or `false`, `per\_evidence` MUST be a verbatim snippet from Next Thinking.\\
- If `per` is `null`, `per\_evidence` MUST be `null`.\\

\#\#\# Rules:\\
1. Keep `obj` and `obj\_evidence` unchanged from the current ledger.\\
2. Update `per` and `per\_evidence` using ONLY the agent’s thinking.\\
\quad - Update `null` to `true` when the agent assumes the constraint holds.\\
\quad - Update `null` to `false` when the agent disqualifies the candidate.\\
\quad - Update existing `true` or `false` ONLY if the agent explicitly states otherwise.\\
3. Add any new Candidates appearing in Next Thinking; new candidates start with all `per=null`.\\
4. Update `status` based on the agent’s intent:\\
\quad - If the agent will answer with X → X is `active`.\\
\quad - If the agent moves away from X without ruling it out → X is `stored`.\\
\quad - If the agent rejects X → X is `rejected`.\\

\#\#\# Output Format (JSON only):\\
\{ "\verb=<=Candidate Name\verb=>=": \{ "status": "active$|$stored$|$rejected", "constraints": \{ "\verb=<=constraint\_id\verb=>=": \{ "obj": do not update, "per": true$|$false$|$null, "obj\_evidence": do not update, "per\_evidence": "\verb=<=verbatim snippet or null\verb=>=" \} \} \} \}\\

\#\#\# Inputs:\\
- **Question:** \{question\}\\
- **Constraints:** \{constraints\}\\
- **Current Perception \& Status Ledger:** \{current\_ledger\}\\
- **Previous Thinking:** \{prev\_thinking\}\\
- **Search Query:** \{query\}\\
- **Search Results:** \{result\}\\
- **Next Thinking:** \{next\_thinking\}\\

\#\#\# Output:\\
Only the updated JSON ledger.\\
\end{tcolorbox}


\paragraph{Prompt for \textsc{LiveLedger} Update from Search Results.}
This prompt instructs the LiveLedger model to analyze newly retrieved search results and update a ledger with objective evidence relevant to candidate answers and their constraints. The model identifies new candidates, determines whether constraints are verified, contradicted, or remain unknown, and records only direct, unambiguous evidence with verbatim citations. By separating evidence extraction from answer generation, this prompt enables reliable, incremental verification across iterative search steps.
\begin{tcolorbox}[breakable, colframe=black, title={Prompt for Ledger Update from Search Results}]
\#\#\# System Instruction:\\
You are a reasoning assistant that updates an epistemic ledger based on search results.\\

\#\#\# Task:\\
Analyze the search results and update the ledger with any new objective evidence found.\\

\#\#\# Input Fields:\\
- **Question:** the multi-constraint question being answered\\
- **Constraints:** labeled conditions (C1, C2, ...) that candidates must satisfy\\
- **Current Ledger:** evidence collected so far\\
- **Latest Step:** the most recent search iteration, containing:\\
\quad - **Thinking:** reasoning that motivated the search\\
\quad - **Search Query:** the query executed\\
\quad - **Search Results:** documents returned by the search\\

\#\#\# Instructions:\\
1. Carefully review the Search Results.\\
2. Identify any candidates not present in the current ledger.\\
3. Identify any evidence that verifies or refutes constraints for candidates.\\
4. Call `update\_ledger` with entries for each piece of new evidence found.\\

\#\#\# Status Guidelines:\\
- `obj = true`: Search Results contain clear, direct evidence that proves the constraint is satisfied.\\
- `obj = false`: Search Results contain clear, direct evidence that disproves the constraint.\\
- `obj = null`: Evidence is ambiguous, indirect, or not found.\\

**Important Rules:**\\
- Include an entry for each candidate not in the current ledger with `obj = null`.\\
- Set `true` or `false` ONLY when evidence is direct and unambiguous.\\
- Always include a verbatim quote in `obj\_evidence` when `obj` is non-null.\\
- If no relevant evidence is found, call `update\_ledger(entries=[])`.\\

\#\#\# Output Requirement:\\
Always call `update\_ledger` based on the Search Results, including:\\
- New candidates with `obj = null`.\\
- Any (candidate, constraint) pair with newly discovered evidence.\\

\#\#\# Tool Call Format:\\
update\_ledger(entries=[\{ "candidate": "\verb=<=Candidate Name\verb=>=", "constraint": "\verb=<=constraint\_id\verb=>=", "obj": true$|$false$|$null, "obj\_evidence": "\verb=<=verbatim quote or null\verb=>=" \}])\\

\#\#\# Inputs:\\
- **Question:** \{question\}\\
- **Constraints:** \{constraints\}\\
- **Current Ledger:** \{ledger\}\\
- **Thinking:** \{thinking\}\\
- **Search Query:** \{search\_query\}\\
- **Search Results:** \{retrieval\_results\}\\

\#\#\# Output:\\
Call `update\_ledger` with your findings.\\
\end{tcolorbox}

\paragraph{Prompt for Answer Correctness Evaluation.}
This prompt instructs the model to determine whether a predicted answer matches a gold reference answer. It enforces strict entity-level equivalence, permitting matches only under unambiguous aliasing while rejecting category mismatches and ambiguous cases. Accuracy is assessed by recall, such that a prediction is considered correct if it matches any valid gold entity. This design ensures consistent and conservative correctness judgments across both single-answer and multi-answer questions.
\begin{tcolorbox}[breakable, colframe=black, title={Prompt for Answer Correctness Evaluation}]
\#\#\# Instruction:\\
You are an impartial evaluator.\\

You will be given:\\
- a **question**\\
- a **gold (reference) answer**, which may contain **one or more valid entities**\\
- a **predicted answer**\\

Your task is to determine whether the **predicted answer is correct**.\\

\#\#\# Rules (follow strictly):\\
1. If the gold answer contains **multiple valid entities**, the predicted answer is correct if and only if it clearly matches **any one** of the gold entities.\\
2. If the gold answer contains a **single entity**, the predicted answer must match **that exact entity**.\\
3. Minor surface differences (e.g., capitalization, abbreviations, aliases, name order) are allowed **only if** they unambiguously refer to the same entity.\\
4. If the predicted answer refers to a **different entity**, the verdict must be **false**, even if it partially satisfies the question.\\
5. If the predicted answer is **more general, more specific, or a different category** than the gold answer, the verdict must be **false**.\\
6. Do **not** use outside knowledge beyond comparing the gold and predicted answers.\\
7. If equivalence or membership is **ambiguous**, default to **false**.\\
8. Do **not** reward partial correctness.\\

\#\#\# Output Format (JSON only):\\
\{ "verdict": true|false, "justification": "One-sentence explanation." \}\\

\#\#\# Inputs:\\
- **Question:** \{question\}\\
- **Gold Answer:** \{answer\}\\
- **Predicted Answer:** \{predicted\_answer\}\\

\#\#\# Output:\\
Only the JSON object.\\
\end{tcolorbox}


\end{document}